\pdfoutput=1
\PassOptionsToPackage{table}{xcolor}
\documentclass[dvipsnames]{article}
\usepackage{colm2024_conference}

\usepackage{amsmath,amsfonts,bm}

\def\eqref#1{equation~\ref{#1}}

\def\1{\bm{1}}

\DeclareMathAlphabet{\mathsfit}{\encodingdefault}{\sfdefault}{m}{sl}
\SetMathAlphabet{\mathsfit}{bold}{\encodingdefault}{\sfdefault}{bx}{n}

\newcommand{\trackone}{ME}
\newcommand{\tracktwo}{RE}
\newcommand{\trackonename}{Multimodal Editing}
\newcommand{\tracktwoname}{Repository Editing}
\newcommand{\trackonefull}{\trackonename{} (\trackone{})}
\newcommand{\tracktwofull}{\tracktwoname{} (\tracktwo{})}
{ME}%
  \newcommand{\tracktwo}{RE}%
  \newcommand{\trackonename}{Multimodal Editing}%
  \newcommand{\tracktwoname}{Repository Editing}%
  \newcommand{\trackonefull}{\trackonename{} (\trackone{})}%
  \newcommand{\tracktwofull}{\tracktwoname{} (\tracktwo{})}%
}

\usepackage{graphicx}
\usepackage{caption}    
\usepackage{placeins}
\usepackage{enumitem}   
\usepackage{marvosym}

\renewenvironment{abstract}{%
  \par\vspace{4pt}\centerline{\large\bfseries Abstract}\vspace{3pt}%
  \small\setlength{\parskip}{0pt}%
}{\par\vspace{4pt}}

\newcommand{\websitelink}{https://codesignbench.github.io/}

\usepackage{xcolor}
\usepackage{colortbl}
\usepackage{booktabs}
\usepackage{array}
\usepackage{multirow}
\usepackage{tikz}
\usepackage{pifont}
\definecolor{cdqwen}  {HTML}{244BB2}   
\definecolor{cdsky}   {HTML}{7090D8}   
\definecolor{cdamber} {HTML}{D1B280}   
\definecolor{cdteal}  {HTML}{4A8D9A}   
\definecolor{cdmauve} {HTML}{6A577D}   
\definecolor{cdslate} {HTML}{A6A7AD}   
\definecolor{cdgreen} {HTML}{789977}   
\definecolor{cdrose}  {HTML}{9B5760}   
\definecolor{cdink}   {HTML}{232326}   

\definecolor{cdwamber}{HTML}{E4AD4B}   
\definecolor{cdwteal} {HTML}{2EA6BA}   
\definecolor{cdwmauve}{HTML}{936CB8}   
\definecolor{cdwqwen} {HTML}{5A8BFC}   

\definecolor{cdband}  {HTML}{EEEFF2}   
\definecolor{cdgrid}  {HTML}{E4E5E9}   
\definecolor{cdfaint} {HTML}{CBCCD4}   
\definecolor{cdmuted} {HTML}{9EA0AC}   

\newcommand{\cdbarw}{1.15}   
\newcommand{\cdbarh}{0.30}   
\newcommand{\cdbarfont}{\fontsize{7.6}{8.4}\selectfont}

\newcommand{\cdwash}{100}
\newcommand{\cdt}[1]{\number\numexpr#1*\cdwash/100\relax}

\newcolumntype{V}{!{\color{cdfaint}\vrule width 0.4pt}}

\newcommand{\cdbar}[4][\cdbarw]{%
  \begin{tikzpicture}[baseline=0.40ex,x=1cm,y=1cm]
    \fill[#2!\cdt{14},rounded corners=1pt] (0,0) rectangle (#1,\cdbarh);
    \shade[left color=#2!\cdt{62},right color=#2!\cdt{24},rounded corners=1pt]
      (0,0) rectangle ({max(#4,0.010)*#1},\cdbarh);
    \node[inner sep=0pt,text=cdink] at ({#1/2},{\cdbarh/2}) {\cdbarfont #3};
  \end{tikzpicture}%
}
\newcommand{\cdbarb}[4][\cdbarw]{%
  \begin{tikzpicture}[baseline=0.40ex,x=1cm,y=1cm]
    \fill[#2!\cdt{14},rounded corners=1pt] (0,0) rectangle (#1,\cdbarh);
    \shade[left color=#2!\cdt{78},right color=#2!\cdt{34},rounded corners=1pt]
      (0,0) rectangle ({max(#4,0.010)*#1},\cdbarh);
    \draw[#2!\cdt{85},rounded corners=1pt,line width=0.45pt] (0,0) rectangle (#1,\cdbarh);
    \node[inner sep=0pt,text=cdink] at ({#1/2},{\cdbarh/2}) {\cdbarfont \bfseries #3};
  \end{tikzpicture}%
}

\newcommand{\cdhm}[3]{\cellcolor{#1!#2}#3}

\newcommand{\cdbest}[1]{\textbf{#1}}
\DeclareRobustCommand{\cdmodel}[1]{{\ttfamily\fontsize{7.3}{8.3}\selectfont #1}}
\newcommand{\cdna}{{\color{cdmuted}--}}

\DeclareMathSizes{5.6}{5.6}{5}{5}
\DeclareRobustCommand{\cdup}{\,{\fontsize{5.6}{6}\selectfont$\uparrow$}}

\newcommand{\cdhead}[1]{\textbf{#1}}
\newcommand{\cdgrouprow}[2]{\rowcolor{cdband}\multicolumn{#1}{@{}l@{}}%
  {\hspace{1.5pt}\textsc{\footnotesize #2}}}

\newcommand{\cdiconbox}[1]{\makebox[0.40cm][c]{\raisebox{-0.20\height}{#1}}}
\DeclareRobustCommand{\cdiconopenai}{\cdiconbox{%
  \includegraphics[width=0.29cm,height=0.29cm,keepaspectratio]{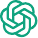}}}
\DeclareRobustCommand{\cdiconkimi}{\cdiconbox{%
  \includegraphics[width=0.29cm,height=0.29cm,keepaspectratio]{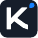}}}

\DeclareRobustCommand{\cdiconanthropic}{\cdiconbox{%
  \includegraphics[width=0.29cm,height=0.29cm,keepaspectratio]{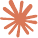}}}
\DeclareRobustCommand{\cdiconqwen}{\cdiconbox{%
  \includegraphics[width=0.29cm,height=0.29cm,keepaspectratio]{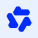}}}
\DeclareRobustCommand{\cdiconglm}{\cdiconbox{%
  \includegraphics[width=0.29cm,height=0.29cm,keepaspectratio]{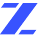}}}
\DeclareRobustCommand{\cdiconxai}{\cdiconbox{%
  \includegraphics[width=0.29cm,height=0.29cm,keepaspectratio]{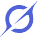}}}
\newcommand{\cdmi}[2]{#1\hspace{0.28em}\cdmodel{#2}}

\newcommand{\auditfix}[1]{#1}

\AddToHook{env/figure/begin}{%
  \setlength{\abovecaptionskip}{4pt}%
  \setlength{\belowcaptionskip}{0pt}}
\AddToHook{env/figure*/begin}{%
  \setlength{\abovecaptionskip}{4pt}%
  \setlength{\belowcaptionskip}{0pt}}
\AddToHook{env/table/begin}{\setlength{\abovecaptionskip}{0pt}}

\hypersetup{
  colorlinks=true,
  breaklinks=true,           
  linkcolor=cdqwen,          
  citecolor=cdqwen!80!cdink, 
  urlcolor=cdqwen,
  pdfborder={0 0 0},
}

\title{%
\texorpdfstring{%
  \raisebox{-0.15em}{%
    \includegraphics[height=1.7em]{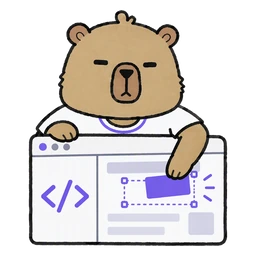}%
  }\hspace{0.3em}%
}{}%
Co-Design Bench:\\ Multimodal Collaborative Coding Evaluation for Agentic Models
}
\author{%
  \makebox[\linewidth][c]{\normalsize\bfseries
  \begin{tabular}{@{}c@{}}
    Xilin Wei\textsuperscript{1,2,$\ast$}\quad
    Shengyuan Ding\textsuperscript{1,$\ast$}\quad
    Xinyu Fang\textsuperscript{2,3,$\ast$}\quad
    Tiancheng Gu\textsuperscript{2}\quad
    Yuchi Wang\textsuperscript{2,4}\quad
    Zhixun Li\textsuperscript{2,4}\\[0.35em]
    Ziyu Liu\textsuperscript{5}\quad
    Yuhang Zang\textsuperscript{6}\quad
    Weixiang Yu\textsuperscript{2}\quad
    Cheng Lu\textsuperscript{2}\quad
    Chonggao Sun\textsuperscript{2}\quad
    Tong Zhou\textsuperscript{2}\\[0.35em]
    Dahua Lin\textsuperscript{4,\textdagger}\quad
    Hang Zhang\textsuperscript{2,\textdagger}\quad
    Shuai Bai\textsuperscript{2,\textdagger}\quad
    Dayiheng Liu\textsuperscript{2}
  \end{tabular}}\\[0.75em]
  \makebox[\linewidth][c]{\small\normalfont
  \begin{tabular}{@{}c@{}}
    \textsuperscript{1}Fudan University\quad
    \textsuperscript{2}Alibaba Token Hub, Alibaba Group\quad
    \textsuperscript{3}Zhejiang University\\
    \textsuperscript{4}The Chinese University of Hong Kong\quad
    \textsuperscript{5}Shanghai Jiao Tong University\quad
    \textsuperscript{6}Nanyang Technological University
  \end{tabular}}\\[0.3em]
    {\small\normalfont
      \textsuperscript{$\ast$}Equal contribution\quad
      \textsuperscript{\textdagger}Corresponding authors
    }\\[0.3em]
    {\small\normalfont
      \href{\websitelink}{\texttt{\websitelink}}
    }%
  }

\hypersetup{
  pdftitle={Co-Design Bench: Multimodal Collaborative Coding Evaluation for Agentic Models},
  pdfauthor={Xilin Wei, Shengyuan Ding, Xinyu Fang, Tiancheng Gu, Yuchi Wang, Zhixun Li, Ziyu Liu, Yuhang Zang, Weixiang Yu, Cheng Lu, Chonggao Sun, Tong Zhou, Dahua Lin, Hang Zhang, Shuai Bai, Dayiheng Liu}
}

\begin{document}

\maketitle

\par\noindent\begin{minipage}{\linewidth}
\centering
\includegraphics[width=\linewidth]{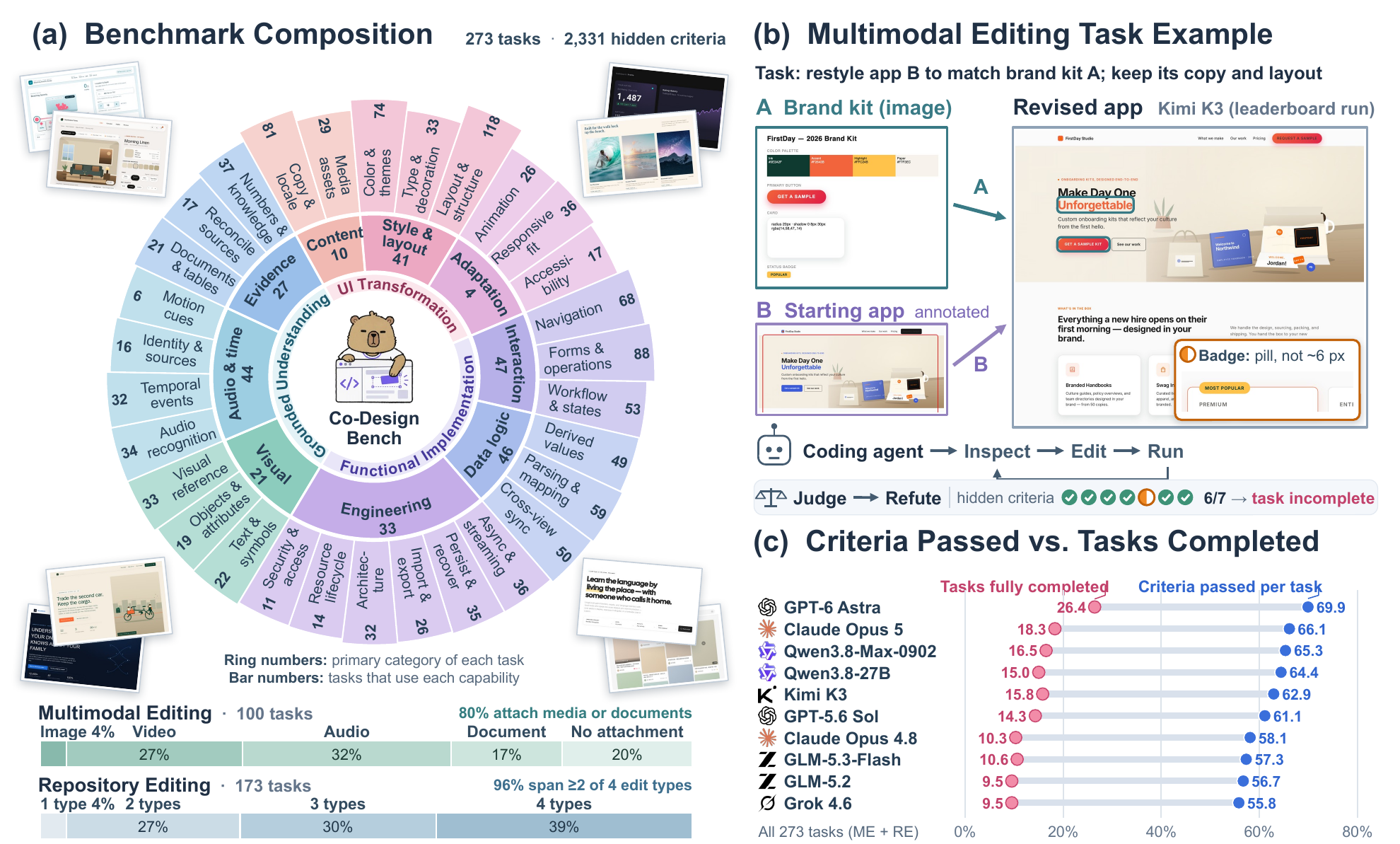}
\captionsetup{font=small,labelfont=bf,skip=4pt,hypcap=false}
\captionof{figure}{\textbf{Co-Design Bench at a glance.}
  \textbf{(a)} $273$ tasks in two tracks, \trackonefull{} and \tracktwofull{}, with $2{,}331$ hidden criteria and a taxonomy of $3$ dimensions, $9$ categories and $30$ capabilities (bar counts are multi-label; each capability spans an equal angle). Cards show starting apps of public \trackone{} tasks; strips show \trackone{} attachment types and how many of four edit types (structure, style, interaction, content) each \tracktwo{} task spans.
  \textbf{(b)} A showcase \trackone{} task from the benchmark: restyle app B to match brand kit A. Kimi K3 passes $6$ of $7$ hidden criteria after refutation, but its badge keeps pill corners instead of the kit's ${\sim}6$\,px, so the task is incomplete.
  \textbf{(c)} Under a shared agentic harness, every endpoint passes over half of the criteria per task on average, yet none fully completes more than $26.4\%$ of tasks.}
\captionsetup{hypcap=true}
\label{fig:teaser}
\end{minipage}\par\vspace{6pt}

\begin{abstract}
Real users ask coding agents to revise existing applications, yet seemingly correct revisions can still leave their requirements unmet.
We introduce \textbf{Co-Design Bench}, a reference-free benchmark of $273$ application-revision tasks curated from $37{,}463$ deduplicated queries across $26{,}314$ real user coding sessions.
Task selection favors difficult, verifiable requests with historical failure signals, including user reformulations and rollbacks.
The tasks retain users' revision goals and available context, pairing runnable starting applications and normalized requests with $2{,}331$ hidden expert-authored acceptance criteria.
Its \trackonefull{} and \tracktwofull{} tracks cover localized edits with optional image, audio, video, or document inputs and coordinated changes across components, state, and data flow.
Our \emph{judge-then-refute} protocol evaluates these criteria using source and browser evidence, challenges positive judgments with a separate verifier, and audits visual and interaction evidence without requiring gold code or target screenshots.
Under a shared agentic harness, the strongest of ten model endpoints passes an average of $69.9\%$ of criteria per task but meets every criterion on only $26.4\%$ of tasks.
Despite similar mean criterion completion, full-pass rates average $23.5\%$ for \trackone{} and $9.5\%$ for \tracktwo{}.
The results show a gap between criterion-level completion and full-task acceptance under the benchmark rubrics for real-user revision requests.
\end{abstract}

\clearpage

\section{Introduction}
\label{sec:intro}

AI coding assistants and agentic coding tools such as Claude Code~\citep{claude_code} and Codex~\citep{codex} have made code generation widely accessible and now support web application co-design, where users inspect an application, request changes, and assess the result \citep{schon1983}.
As these tools reach everyday users, the open question shifts from whether models can write code~\citep{qwen37plus,qwen38} to whether they deliver what users actually ask for in their own applications.
Real coding sessions show where this falls short: some requests are reformulated again and again or end in a rollback.
Such requests can combine image~\citep{internlmxcomposer2,liu2024mmdu}, audio, video~\citep{chen2024sharegpt4video,wei2025videorope,internlmxcomposer2_5_OL}, or document inputs with coordinated changes to layout, state, and application logic, and a revision can look right while missing one requirement.
In Figure~\ref{fig:teaser}b, Kimi~K3 restyles an application to match an attached brand kit and passes $6$ of $7$ hidden criteria, but a badge keeps pill corners instead of the kit's ${\sim}6$\,px, so the task is incomplete.

Existing benchmarks~\citep{refcraft} capture parts of this problem, from reference reconstruction \citep{design2code, webrenderbench, iwrbench} and synthetic or predefined edits \citep{mtweb2code, webcompass} to version-history edits \citep{designbench} and repository issues or freelance requests \citep{swebench, swebenchmm, swelancer}, yet few derive tasks from real coding sessions, where such hard cases surface (Table~\ref{tab:related}).
Live pairwise human preference \citep{chatbotarena}, as in WebDev Arena \citep{arenawebdev}, does rank models on real requests, but each vote records only a preference between two outputs, not which requirement was missed.
We therefore mine real coding sessions, favoring requests that users had to reformulate or roll back, and evaluate each task against expert-authored acceptance criteria, reporting both criterion satisfaction and complete task success.
Identifying missed criteria and measuring full-task acceptance reveal where coding assistants fall short on real users' revision requests.


We introduce \textbf{Co-Design Bench} (Figure~\ref{fig:teaser}a), comprising $273$ tasks and $2{,}331$ expert-authored acceptance criteria, curated from $37{,}463$ deduplicated queries across $26{,}314$ real user coding sessions.
We recover and validate runnable pre-edit applications, retain available session context and attachments, and consolidate user requests into bounded editing tasks.
Candidates are ranked by structural difficulty, verifiability, and historical failure signals such as repeated reformulations and rollbacks.
The benchmark covers two complementary settings: \emph{\trackonefull{}} evaluates localized visual, structural, and content edits with optional image, audio, video, or document inputs, attached in $80\%$ of its $100$ tasks; \emph{\tracktwofull{}} evaluates coordinated UI and feature changes across components, routes, state, and data flow, with $96\%$ of its $173$ tasks spanning at least two of four edit types.
Tasks are organized into $3$ capability dimensions (Grounded Understanding, UI Transformation, and Functional Implementation), $9$ categories, and $30$ capabilities.
Both tracks pair starting applications with normalized requests and hidden criteria specifying requested changes, usually allowing multiple valid implementations.

Evaluating these tasks requires checking whether requested features are reachable and behave correctly in the edited application.
Our \emph{judge-then-refute} protocol (Figure~\ref{fig:teaser}b) first grades hidden criteria using source code and browser execution evidence.
A separate LLM verifier adversarially challenges positive verdicts against the collected evidence, checking for stubs, hardcoding, unreachable functionality, and requirement mismatches.
Audits require successful renders and screenshots for visual claims and recorded browser actions for interaction claims; missing evidence aborts evaluation before score aggregation.
The pre-edit application grounds the requested changes, enabling reference-free acceptance without gold code or target screenshots, while retained verdicts and traces link each criterion judgment to its supporting evidence for inspection.

We evaluate ten model endpoints under a shared agentic harness.
The strongest endpoint passes an average of $69.9\%$ of criteria per task but fully satisfies only $26.4\%$ of tasks; every endpoint shows this pattern (Figure~\ref{fig:teaser}c).
Although per-task criterion pass rates are similar on the two tracks ($62.6\%$ on \trackone{}, $61.3\%$ on \tracktwo{}), every endpoint fully satisfies a smaller share of \tracktwo{} than of \trackone{} tasks ($9.5\%$ vs.\ $23.5\%$ on average), a gap that persists at matched rubric size.
These results motivate measuring both criterion-level completion and full-task acceptance, since satisfying many criteria can still leave a revision incomplete.

\noindent\textbf{Contributions.}
\textbf{(1)}~We introduce Co-Design Bench: $273$ application-revision tasks derived from real user coding sessions, with runnable starting applications and retained context for localized multimodal edits and coordinated repository changes.
\textbf{(2)}~We develop judge-then-refute, a reference-free protocol that combines source and browser evidence, adversarial verification, and mandatory audits to assess the edited application against hidden acceptance criteria.
\textbf{(3)}~We characterize the gap between criterion-level completion and full-task acceptance across ten model endpoints through modality, capability, and failure analyses.

\section{Related Work}
\label{sec:related}

\begin{table}[t]
\centering
\caption{\textbf{Comparison of web-coding benchmarks.}
Task origins and core evaluation capabilities of the reported settings.}
\label{tab:related}

\begingroup
\hypersetup{hidelinks}
\colorlet{tableyes}{cdgreen!70!cdink}
\colorlet{tableno}{cdmuted!65!cdink}
\colorlet{tablepartial}{cdamber!65!cdink}
\colorlet{tablecite}{cdink!72}
\newcommand{\cmark}{\textcolor{tableyes}{\ensuremath{\checkmark}}}
\newcommand{\xmark}{\textcolor{tableno}{\ensuremath{\times}}}
\newcommand{\pmark}{\textcolor{tablepartial}{\ensuremath{\triangle}}}
\newcommand{\benchcite}[2]{\shortstack[l]{#1\\{\scriptsize
  \textcolor{tablecite}{\citep{#2}}}}}
\newcommand{\thstack}[1]{{\bfseries
  \shortstack[c]{#1}}}
\small
\setlength{\tabcolsep}{1.2pt}
\renewcommand{\arraystretch}{0.96}
\begin{tabular*}{\linewidth}{@{\extracolsep{\fill}}l c c c c c c@{}}
\toprule
\textbf{Benchmark / subset}
  & \multicolumn{3}{c}{\textbf{Task}}
  & \multicolumn{3}{c}{\textbf{Evaluation}} \\
\cmidrule(lr){2-4}\cmidrule(lr){5-7}
  & \thstack{Setting}
  & \thstack{Session-derived\\edits}
  & \thstack{Input\\modalities}
  & \thstack{Reference-free\\acceptance}
  & \thstack{Agentic\\evaluation}
  & \thstack{Adversarial\\verification} \\
\midrule
\cdgrouprow{7}{Reference reconstruction} \\
\benchcite{Design2Code}{design2code}
  & Recon. & \xmark & I & \xmark & \xmark & \xmark \\
\benchcite{WebSight-Test}{waffle}
  & Recon. & \xmark & I & \xmark & \xmark & \xmark \\
\benchcite{WebRenderBench}{webrenderbench}
  & Recon. & \xmark & I & \xmark & \xmark & \xmark \\
\benchcite{IWR-Bench}{iwrbench}
  & Recon. & \xmark & I,V & \xmark & \xmark & \xmark \\
\cdgrouprow{7}{Page-level coding} \\
\benchcite{MT-Web2Code}{mtweb2code}
  & Recon./Repair & \xmark & T,I & \xmark & \xmark & \xmark \\
\benchcite{DesignBench}{designbench}
  & Gen./Edit/Repair & \pmark & T,I & \pmark & \xmark & \xmark \\
\benchcite{WebCompass (edit)}{webcompass}
  & Edit & \xmark & T,I & \cmark & \xmark & \xmark \\
\benchcite{MiniAppBench}{miniappbench}
  & Gen. & \xmark & T & \cmark & \cmark & \xmark \\
\cdgrouprow{7}{Repository software engineering} \\
\benchcite{SWE-bench M}{swebenchmm}
  & Repair & \xmark & T,I,V & \pmark & \xmark & \xmark \\
\benchcite{SWE-Lancer (IC)}{swelancer}
  & Edit & \xmark & T & \cmark & \xmark & \xmark \\
\rlap{\color{cdqwen!3}\rule[-3.5pt]{\linewidth}{15.5pt}}%
\textbf{Co-Design Bench}\,{\color{tablecite}(ours)}
  & Edit & \cmark & T,\,I,\,V,\,A,\,D & \cmark & \cmark & \cmark \\
\bottomrule
\end{tabular*}

\vspace{2pt}
\begin{minipage}{\linewidth}
\scriptsize
\cmark~=reported; \xmark~=not reported; \pmark~=subset/metric-dependent.
Gen.~=~generation from scratch; Recon.~=~reference reconstruction; Edit/Repair revise existing code.
Session-derived edits come from user coding histories, excluding issue reports and standalone generation prompts;
DesignBench's session-derived requests belong to its editing subset.
MiniAppBench has 77 direct real-user queries and 423 LLM-augmented variants.
T/I/V/A/D~=text/image/video/audio/document.
Reference-free acceptance needs no gold code or target screenshot; pre-edit baselines are allowed.
Agentic evaluation chooses tools/actions beyond fixed replay. Adversarial verification separately challenges positive verdicts.
\end{minipage}
\endgroup
\end{table}

\noindent\textbf{Multimodal web generation and editing.}
Design2Code~\citep{design2code}, WebSight-Test (from WAFFLE~\citep{waffle}), and WebRenderBench~\citep{webrenderbench} assess screenshot-to-code reconstruction against reference pages. IWR-Bench adds interaction videos with annotated action replay and reference screenshots \citep{iwrbench}. WebGen-Bench and FrontendBench use browser or sandbox functional tests for generation from scratch \citep{webgenbench, frontendbench}. MiniAppBench evaluates interactive application generation as single self-contained HTML files from text prompts \citep{miniappbench}. For editing, DesignBench extracts adjacent versions from V0/Vue0 histories for before/after screenshot assessment \citep{designbench}; MT-Web2Code corrupts reference pages to synthesize repairs \citep{mtweb2code}, and WebCompass uses predefined edit operations \citep{webcompass}. FronTalk studies conversational generation with multimodal feedback \citep{frontalk}. Co-Design Bench pairs runnable baselines with session-derived multimodal and repository revision requests, using hidden criteria without gold code or target screenshots.

\noindent\textbf{Repository-level coding.}
SWE-bench tests real GitHub issue fixes \citep{swebench}; SWE-bench Multimodal adds visual tests for frontend JavaScript \citep{swebenchmm}, and SWE-Lancer uses browser end-to-end tests for real freelance requests \citep{swelancer}~\citep[cf.][]{wildclawbench,nonmarkovgames}. Co-Design Bench's repository track evaluates compound design and feature revisions across components, routes, state, and data flow. Both tracks share a task interface and acceptance protocol measuring requirement coverage and complete delivery.

\noindent\textbf{Agentic artifact evaluation and verification.}
Agent-as-a-Judge evaluates artifacts against user requirements \citep{agentjudge}~\citep[cf.][]{duan2024vlmevalkit,blenderworld}. MiniAppBench's main reference-free evaluator autonomously uses Playwright to assess intention, static implementation, and dynamic behavior from DOM and source evidence, explicitly excluding screenshots \citep{miniappbench}. WebCompass uses adaptive browser tests for generation and non-interactive checklist judgments over code, logs, and screenshots for editing and repair \citep{webcompass}. Structured LLM judges can improve human agreement but can exhibit self-preference and benchmark-dependent reliability \citep{geval, selfpref, agentrewardbench}. Our \emph{judge-then-refute} uses source and browser evidence with a separate verifier: source presence alone cannot establish reachability or correct behavior. Mandatory audits require screenshots for visual claims and action logs for interactions; retained evidence supports verdict inspection but cannot eliminate judgment error in automated evaluation.

\begin{figure}[t]
\centering
\includegraphics[width=\textwidth]{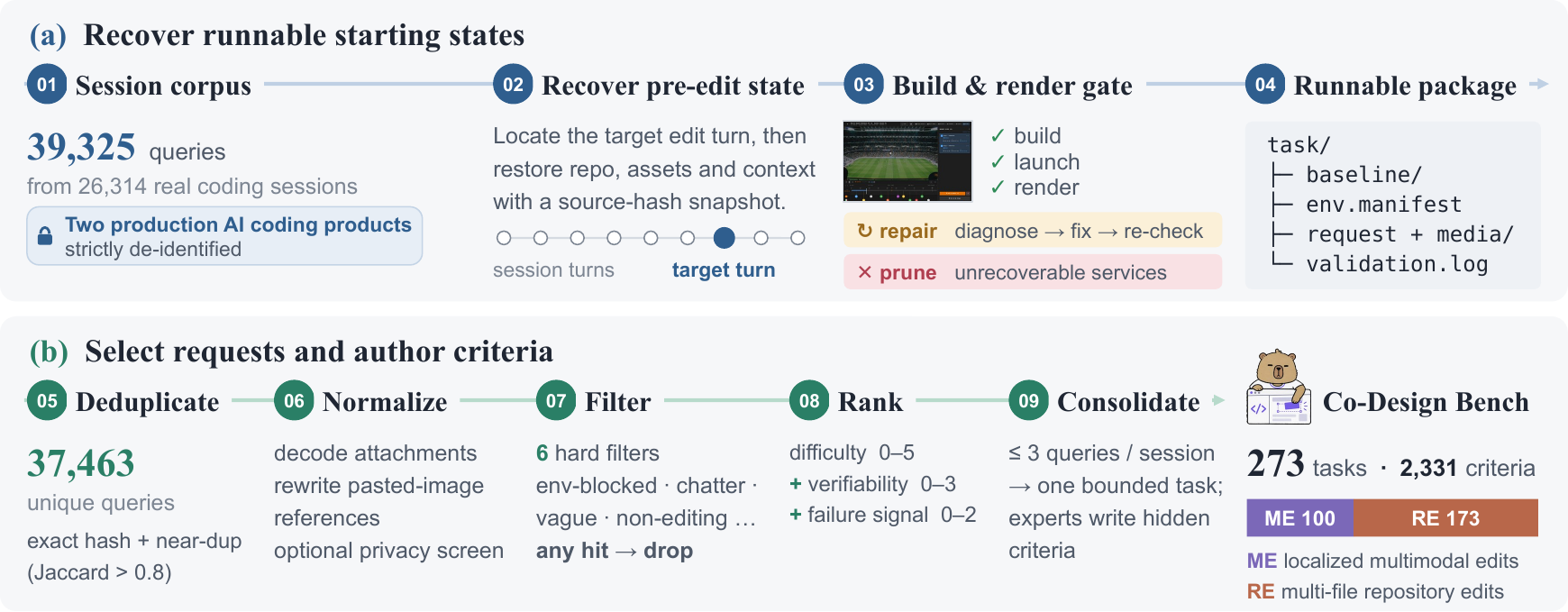}
\caption{\textbf{Data construction and filtering.} (a) Recover and validate runnable pre-edit applications, repairing eligible environments and excluding unrecoverable cases. (b) Deduplicate and filter requests, rank candidates by difficulty, verifiability, and historical failure signals, then select bounded session intervals for consolidation and expert rubric authorship.}
\label{fig:curation_pipeline}
\end{figure}

\section{Co-Design Bench}
\label{sec:benchmark}

Co-Design Bench evaluates whether coding agents can revise existing applications to meet real users' requirements. It contains $273$ tasks and $2{,}331$ expert-authored acceptance criteria. Each task pairs a runnable starting application with a fixed revision request; hidden criteria assess the edited result. The benchmark covers two edit scopes: localized changes in \trackonefull{} and coordinated changes across a repository in \tracktwofull{}.

\subsection{Task Formulation}
\label{sec:benchmark:formulation}

We represent a task as $t=(S,I,C)$. The starting state $S=(B,E)$ contains the pre-edit application $B$ and its containerized environment $E$. The input $I=(q,X,K)$ contains a normalized request $q$, optional multimodal attachments $X$, and retained session history $K$. Through a solver harness $H$, the agent receives $(S,I)$ and can inspect, edit, build, and run the application without further user feedback.

The rubric $C=\{c_i\}_{i=1}^{n_t}$ contains expert-authored acceptance criteria and remains hidden, together with evaluation artifacts. Criteria cover requested outcomes, explicit preservation requirements, and task-specific architectural or implementation constraints. Acceptance concerns the final application state. The pre-edit application provides context for the requested changes and any explicit preservation requirements; reference-free acceptance does not require gold code or a target screenshot. Both tracks use \textsc{Pass}/\textsc{Partial}/\textsc{Fail} verdicts, with completion metrics defined in Section~\ref{sec:eval}.

\subsection{Data Source and Curation}
\label{sec:benchmark:source}

The source corpus contains $39{,}325$ production queries from $26{,}314$ real user coding sessions. Deduplication leaves $37{,}463$ candidates, each linking a target edit turn to its repository, available history, and attachments. Figure~\ref{fig:curation_pipeline} summarizes how these records become executable editing tasks.

\noindent\textbf{Recover runnable applications.}
We reconstruct pre-edit applications from session logs, repository archives, or HTML bundles. Each must pass dependency, build, container-launch, and Playwright rendering checks~\citep{playwright}. Repaired applications pass the same checks; cases requiring unrecoverable services are excluded from the task pool.

\noindent\textbf{Select actionable requests.}
We package attachments and normalize request wording, then remove environment-blocked, non-actionable, and non-editing requests. Eligible candidates are ranked by structural difficulty, verifiability, and \textbf{historical failure signals}, including repeated reformulations and rollbacks. This selection favors challenging requests whose outcomes can be checked. Reformulations and rollbacks are historical failure signals, rather than direct measures of user satisfaction in the fixed-task evaluation.

\noindent\textbf{Consolidate requests and author criteria.}
The selector retains at most three eligible queries per session. Their first and last message indices bound an interval that also includes intervening user turns in chronological order. Experts check the consolidated request and author hidden acceptance criteria. The resulting tasks retain the user's revision goal, available context, and a runnable starting application.

\begin{figure}[t]
\centering
\includegraphics[width=\linewidth]{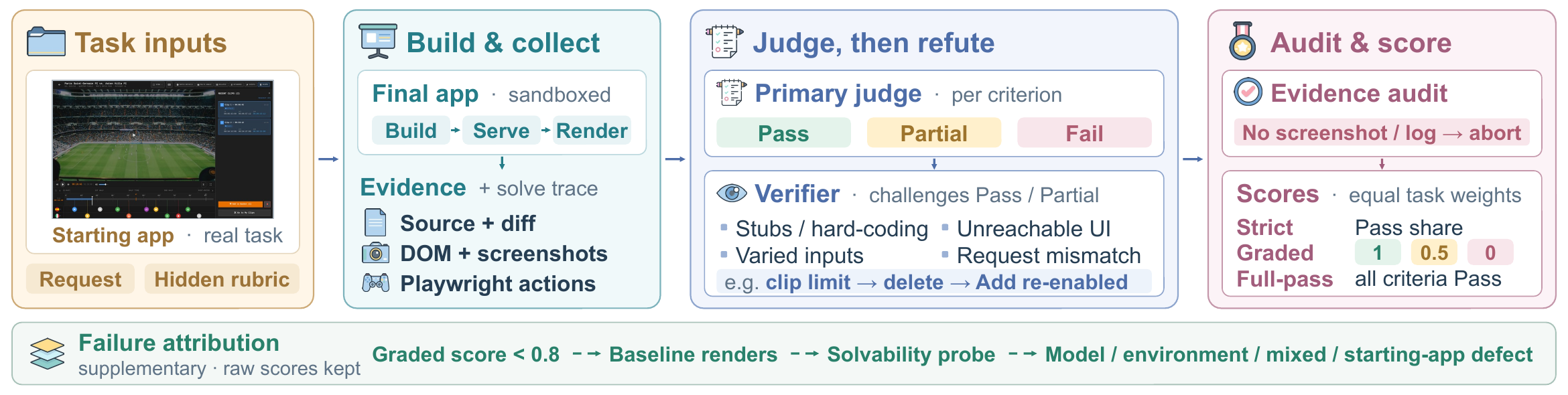}
\caption{\textbf{Evaluation from artifact to verdict.} Build and render the candidate, collect source and browser evidence, and grade hidden criteria. A separate verification pass challenges positive verdicts before evidence audits and task-normalized scoring. The inset shows a real starting app; the example check is derived from its rubric. Parallel baseline and solvability checks support supplementary failure attribution.}
\label{fig:eval_protocol_flow}
\end{figure}

\subsection{\trackonefull}
\label{sec:benchmark:track1}

The $100$ \trackone{} tasks require localized visual, structural, or content edits. Each provides a runnable page, an annotated target region, and a normalized request; $80$ tasks also include image, audio, video, or document attachments for direct or tool-based inspection. Validation checks attachment decodability, target-region annotations, and criterion verifiability in addition to application startability.

Rubrics specify visual treatments, content placement, spatial relationships, and explicit preservation requirements while allowing different CSS/DOM implementations. For example, a receipt-grounded task requires a value to agree across a source document, an editable row, and dependent summaries. Its $553$ acceptance criteria average $5.5$ per task.

\subsection{\tracktwofull}
\label{sec:benchmark:track2}

The $173$ \tracktwo{} tasks require coordinated changes across components, routes, state, data flow, or content. Each provides a runnable multi-file frontend repository, a consolidated request, and local assets. Validation adds route-reachability checks to application startability. The distinction between tracks concerns the scope of the requested edit: a localized \trackone{} revision can also start from a bundled application.

Rubrics assess the requested changes through separately auditable criteria. For example, a clip editor must enforce duration and count limits and recover after clips are removed or split, requiring the interface and application state to stay consistent. This track contains $1{,}778$ criteria, averaging $10.3$ per task. Of its $173$ tasks, $166$ combine multiple structural, stylistic, interactive, or content edit types.

Explicit wiring and artifact constraints are retained; other implementation choices generally remain open. Some \tracktwo{} criteria nevertheless require selectors or class names absent from the request. This limits how directly their verdicts measure satisfaction of the stated user requirements; Section~\ref{sec:exp:failures} analyzes these cases.

The appendix details curation (Appendix~\ref*{sec:app:curation}), task composition and examples (Appendix~\ref*{sec:app:composition} and \ref*{sec:app:cases}), modality, edit-type, and rubric-size statistics (Appendix~Table~\ref*{tab:dataset_stats}), and failure analysis (Appendix~\ref*{sec:app:failure_types}). Table~\ref{tab:related} compares Co-Design Bench with related benchmarks.

\section{Evaluation Protocol}
\label{sec:eval}

Evaluation assesses the final application $W$ against the request and hidden criteria, using baseline $B$ to ground the changes. Figure~\ref{fig:eval_protocol_flow} shows the \emph{judge-then-refute} pipeline.

\subsection{Evidence and Adversarial Verification}
\label{sec:eval:evidence}

\noindent\textbf{Evidence and primary grading.}
The candidate is built and rendered in an isolated environment. A fixed agentic evaluator \citep{agentjudge} assigns \textsc{Pass}, \textsc{Partial}, or \textsc{Fail} to each criterion using source diffs, rendered pages, and Playwright traces. Evidence must match the claim: source inspection can establish a static property, a screenshot can support a visual requirement, and recorded actions can establish an interaction and its resulting state. Failed builds or missing DOM nodes invalidate render-dependent criteria; source presence alone cannot establish that the requested functionality is reachable and usable through the rendered interface.

\noindent\textbf{Refutation and evidence audits.}
A separate LLM verifier receives the initial judgments and collected evidence, then challenges \textsc{Pass} and \textsc{Partial} verdicts. It checks for stubs, hardcoded outputs, unreachable entry points, failures under varied inputs, and request mismatches. This pass targets unsupported completion credit; it does not by itself establish the correctness of every verdict. Subsequent audits require successful renders and screenshots for visual claims and browser-action logs for interaction claims. Violations abort evaluation before aggregation. Verdicts, evidence, and refutation logs are retained for inspection, linking each reported outcome to its supporting observations. Appendix~\ref*{sec:app:eval_implementation} specifies execution, evidence extraction, and artifact retention.

\noindent\textbf{Human--evaluator agreement.}
We compare Opus 4.8's initial and post-refutation judgments with reference labels checked by frontend experts on $100$ saved tasks: $50$ \trackone{} tasks and $50$ \tracktwo{} tasks. Both stages use the same outputs and reference labels. Table~\ref{tab:judge_proxy_agreement} reports human--evaluator exact agreement and linear-weighted Cohen's $\kappa$. Appendix~\ref*{sec:app:judge_proxy} provides further details of the comparison procedure.

\subsection{Scoring and Failure Attribution}
\label{sec:eval:scoring}

For task $t$ with $n_t$ nonempty implementation criteria, let $v_{ti}$ denote the final verdict after refutation and auditing, and let $e_t$ and $p_t$ be the evaluator's recorded earned and maximum points. Strict and Graded Completion are
\begin{equation}
  R_t = \frac{1}{n_t}\sum_{i=1}^{n_t}\mathbf{1}[v_{ti}=\textsc{Pass}],
  \quad
  G_t = \frac{e_t}{p_t},
  \label{eq:task_completion}
\end{equation}
where $\mathbf{1}[\cdot]$ is the indicator function. The evaluator awards full, half, or no credit for \textsc{Pass}, \textsc{Partial}, or \textsc{Fail}, respectively; Graded Completion retains the recorded task point totals. Failed evaluations contribute zero. Over a fixed task set $\mathcal{T}$ shared by all endpoints, we report equal-weight task means and Full-Pass Rate:
\begin{equation}
  \begin{gathered}
  \bar R = \frac{\sum_{t\in\mathcal{T}}R_t}{|\mathcal{T}|},
  \quad
  \bar G = \frac{\sum_{t\in\mathcal{T}}G_t}{|\mathcal{T}|},
  \\
  \mathrm{F} = \frac{\sum_{t\in\mathcal{T}}\mathbf{1}[R_t=1]}{|\mathcal{T}|}.
  \end{gathered}
  \label{eq:completion_aggregation}
\end{equation}
Task normalization prevents longer rubrics from dominating the mean. Strict Completion measures fully satisfied criteria, Graded Completion credits partial progress, and Full-Pass Rate measures whether every criterion in a task is met. All three metrics lie in $[0,1]$ and are reported on a $0$--$100$ scale, with $\mathrm{F}\le\bar R$; the gap $\bar R-\mathrm{F}$ is criterion credit from tasks with unfinished requirements. A high $\bar R$ can therefore coexist with a low $\mathrm{F}$ when unresolved requirements are spread across many tasks. Combined scores weight all $273$ tasks equally: $100$ on \trackone{} and $173$ on \tracktwo{} for each endpoint.

\noindent\textbf{Failure attribution.}
For low-scoring cases ($G_t<0.8$), attribution proceeds when an independently rendered baseline passes the startability gate. A parallel solvability probe attempts a minimal patch in an isolated sandbox, stubbing inaccessible network endpoints while preserving core business logic. The checks distinguish model limitations, environmental blockage, mixed causes, and defects in the starting artifact. Raw scores and denominators remain unchanged. Supplementary adjusted scores grant attribution credit only when infrastructure failures are established (Appendix~\ref*{sec:app:failure_attribution}).

\section{Experiments}
\label{sec:exp}

\subsection{Experimental Setup}
\label{sec:exp:setup}

We evaluate ten frontier models---GPT-6~Astra~\citep{gpt6astra}, Claude Opus~5~\citep{claude_opus5}, Qwen3.8-Max~\citep{qwen38max,qwen38max0902}, Qwen3.8-27B~\citep{qwen38_27b}, Kimi~K3~\citep{kimik3}, GPT-5.6~Sol~\citep{gpt56sol}, Claude Opus~4.8~\citep{claude_opus48}, GLM-5.3-Flash~\citep{glm53flash}, GLM-5.2~\citep{glm52}, and Grok~4.6~\citep{grok46}---as coding agents on all $273$ tasks of Co-Design Bench. All models run in the same Claude Code harness~\citep{claude_code} with low reasoning effort, a maximum of $120$ turns, and a three-hour limit per task. Each agent receives the request, its attachments, and a clean copy of the starting application; the criteria stay hidden. Claude Opus~4.8 serves as both judge and refuter. We report strict completion, graded completion, and full-pass rate, together with median turns, output tokens per turn, and token usage per task. Further details are provided in Appendix~\ref*{sec:app:implementation}.

\subsection{Main Results}
\label{sec:exp:main}

\begin{table}[t]
\centering
\caption{\textbf{Main benchmark performance on \trackone{} and \tracktwo{}.} Models are ranked by overall mean strict completion $\bar R$ on all $273$ tasks; bold marks the best completion values. Strict ($\bar R$) and Graded ($\bar G$) are on a $0$--$100$ scale, and Full is the full-pass rate ($\mathrm{F}$, in \%). Rollout statistics summarize median turns, mean turn output tokens (Len.), and mean total token usage per scored case (Tokens).}
\label{tab:leaderboards}
\fontsize{7.8}{9.2}\selectfont
\renewcommand{\cdmodel}[1]{{\fontencoding{OT1}\ttfamily\fontsize{7.8}{9.2}\selectfont #1}}
\renewcommand{\cdbarfont}{\fontsize{7.8}{9.2}\selectfont}
\setlength{\tabcolsep}{2pt}
\renewcommand{\arraystretch}{1.1}
\renewcommand{\cdbarw}{0.98}
\renewcommand{\cdwash}{75}
\begin{tabular}{@{}l rrr@{\hspace{5pt}} rrr@{\hspace{5pt}} crr V rrc@{\hspace{5pt}} rrc@{}}
\toprule
\multirow{3}{*}{\cdhead{Model}} & \multicolumn{9}{cV}{\cdhead{Completion}} & \multicolumn{6}{c}{\cdhead{Rollout statistics}} \\
\cmidrule(lr){2-10}\cmidrule(l){11-16}
 & \multicolumn{3}{c@{\hspace{5pt}}}{\trackone{}\;{\normalfont\fontsize{7.5}{8.5}\selectfont$n{=}100$}} & \multicolumn{3}{c@{\hspace{5pt}}}{\tracktwo{}\;{\normalfont\fontsize{7.5}{8.5}\selectfont$n{=}173$}} & \multicolumn{3}{cV}{Overall\;{\normalfont\fontsize{7.5}{8.5}\selectfont$n{=}273$}} & \multicolumn{3}{c@{\hspace{5pt}}}{\trackone{}} & \multicolumn{3}{c}{\tracktwo{}} \\
\cmidrule(lr){2-4}\cmidrule(lr){5-7}\cmidrule(lr){8-10}\cmidrule(lr){11-13}\cmidrule(l){14-16}
 & Strict & Graded & Full & Strict & Graded & Full & Strict & Graded & Full & Turns & Len. & Tokens & Turns & Len. & Tokens \\
\midrule
\cdmi{\cdiconopenai}{gpt-6-astra} & \textbf{75.4} & \textbf{79.8} & \textbf{41.0} & \textbf{66.7} & \textbf{72.8} & \textbf{17.9} & \cdbarb{cdwqwen}{69.9}{1.0000} & \textbf{75.3} & \textbf{26.4} & 32 & 146 & \cdbar[0.84]{cdwamber}{1.24M}{0.3968} & 48 & 272 & \cdbar[0.84]{cdwamber}{2.33M}{0.6115} \\
\cdmi{\cdiconanthropic}{opus-5} & 69.5 & 74.9 & 30.0 & 64.2 & 70.2 & 11.6 & \cdbar{cdwqwen}{66.1}{0.9462} & 71.9 & 18.3 & 34 & 363 & \cdbar[0.84]{cdwamber}{2.92M}{0.6891} & 35 & 697 & \cdbar[0.84]{cdwamber}{3.31M}{0.7310} \\
\cdmi{\cdiconqwen}{qwen3.8-max-0902} & 66.7 & 71.9 & 31.0 & 64.4 & 71.5 & 8.1 & \cdbar{cdwqwen}{65.3}{0.9340} & 71.6 & 16.5 & 42 & 620 & \cdbar[0.84]{cdwamber}{2.94M}{0.6906} & 45 & 893 & \cdbar[0.84]{cdwamber}{4.01M}{0.7969} \\
\cdmi{\cdiconqwen}{qwen3.8-27b} & 64.2 & 70.5 & 23.0 & 64.5 & 71.1 & 10.4 & \cdbar{cdwqwen}{64.4}{0.9218} & 70.9 & 15.0 & 56 & 845 & \cdbar[0.84]{cdwamber}{5.90M}{0.9277} & 57 & 1185 & \cdbar[0.84]{cdwamber}{6.72M}{0.9720} \\
\cdmi{\cdiconkimi}{kimi-k3} & 66.2 & 71.8 & 26.0 & 61.0 & 68.1 & 9.8 & \cdbar{cdwqwen}{62.9}{0.8998} & 69.5 & 15.8 & 27 & 329 & \cdbar[0.84]{cdwamber}{1.47M}{0.4546} & 24 & 534 & \cdbar[0.84]{cdwamber}{1.36M}{0.4278} \\
\cdmi{\cdiconopenai}{gpt-5.6-sol} & 58.3 & 63.9 & 20.0 & 62.7 & 70.2 & 11.0 & \cdbar{cdwqwen}{61.1}{0.8742} & 67.9 & 14.3 & 32 & 208 & \cdbar[0.84]{cdwamber}{1.54M}{0.4716} & 40 & 262 & \cdbar[0.84]{cdwamber}{1.80M}{0.5238} \\
\cdmi{\cdiconanthropic}{opus-4-8} & 59.3 & 65.7 & 16.0 & 57.5 & 64.5 & 6.9 & \cdbar{cdwqwen}{58.1}{0.8321} & 64.9 & 10.3 & 24 & 560 & \cdbar[0.84]{cdwamber}{1.65M}{0.4954} & 23 & 815 & \cdbar[0.84]{cdwamber}{1.77M}{0.5176} \\
\cdmi{\cdiconglm}{glm-5.3-flash} & 53.8 & 59.9 & 18.0 & 59.4 & 66.2 & 6.4 & \cdbar{cdwqwen}{57.3}{0.8207} & 63.9 & 10.6 & 49 & 357 & \cdbar[0.84]{cdwamber}{3.65M}{0.7643} & 31 & 481 & \cdbar[0.84]{cdwamber}{1.86M}{0.5354} \\
\cdmi{\cdiconglm}{glm-5.2} & 57.2 & 62.9 & 16.0 & 56.4 & 64.1 & 5.8 & \cdbar{cdwqwen}{56.7}{0.8112} & 63.7 & 9.5 & 69 & 659 & \cdbar[0.84]{cdwamber}{7.29M}{1.0000} & 53 & 603 & \cdbar[0.84]{cdwamber}{3.28M}{0.7280} \\
\cdmi{\cdiconxai}{grok-4.6} & 55.2 & 60.0 & 14.0 & 56.2 & 63.0 & 6.9 & \cdbar{cdwqwen}{55.8}{0.7992} & 61.9 & 9.5 & 15 & 423 & \cdbar[0.84]{cdwamber}{0.65M}{0.1800} & 9 & 1135 & \cdbar[0.84]{cdwamber}{0.44M}{0.0475} \\
\bottomrule
\end{tabular}
\end{table}

Table~\ref{tab:leaderboards} reports completion and rollout statistics, from which we make the following observations.

\begin{figure}[t]
\begin{minipage}[t]{0.52\textwidth}
\centering
\includegraphics[width=\linewidth]{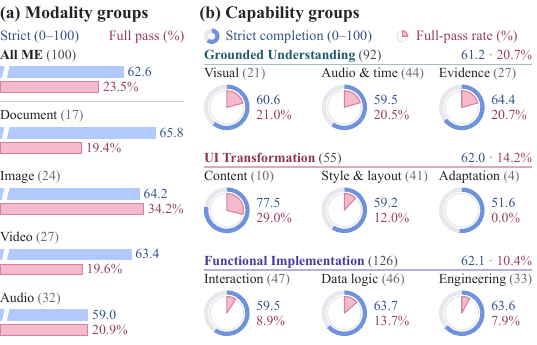}
\caption{Performance by (a)~modality groups and (b)~capability groups, averaged equally across models. Image in (a) includes tasks with image attachments and those without additional attachments.}
\label{fig:sec5_task_breakdown}
\end{minipage}\hfill
\begin{minipage}[t]{0.455\textwidth}
\centering
\includegraphics[width=\linewidth]{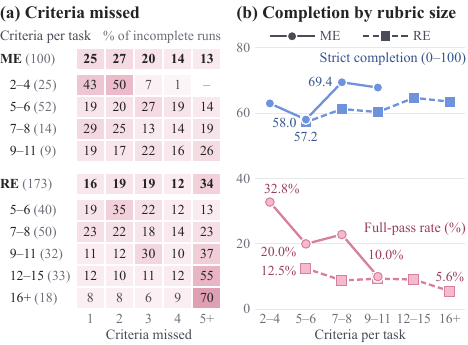}
\caption{Completion gap by rubric size: (a)~criteria missed per incomplete run and (b)~strict completion and full-pass rate, using the archived ten-model cohort.}
\label{fig:completion_gap}
\end{minipage}
\end{figure}

\noindent\textbf{Multimodal collaborative coding remains far from solved.}
Overall strict completion ranges from $55.8$ to $69.9$, yet even the leader, GPT-6~Astra, fully passes only $26.4\%$ of tasks, and no other model more than $18.3\%$: models satisfy most individual criteria but rarely deliver a revision that meets every criterion. The gap widens on repository edits. Mean strict completion is similar on the two tracks ($62.6$ on \trackone{}, $61.3$ on \tracktwo{}), but every model's full-pass rate is lower on \tracktwo{} (mean $23.5\%$ vs.\ $9.5\%$); \tracktwo{} rubrics are about twice as long, but the gap persists at equal rubric size (Section~\ref{sec:exp:gap}).

\noindent\textbf{Excelling at both multimodal grounding and repository coordination remains challenging.}
Strength on one track only partly predicts strength on the other (Spearman $\rho=0.78$ between the track rankings). Kimi~K3 outscores GPT-5.6~Sol by $7.9$ points on \trackone{} but trails it by $1.7$ on \tracktwo{}; only \mbox{GPT-6}~Astra leads both. These results highlight the challenge of achieving strong performance in both capabilities, motivating joint consideration in Co-Design Bench.

\noindent\textbf{More budget does not necessarily yield better performance.}
Opus~5, Qwen3.8-Max, and Qwen3.8-27B use more tokens on average than GPT-6~Astra, yet achieve lower completion on both tracks. For example, Qwen3.8-27B averages $5.90$M tokens per task on \trackone{} and $6.72$M on \tracktwo{}, compared with $1.24$M and $2.33$M for GPT-6~Astra, while scoring $11.2$ and $2.2$ points lower in strict completion. Higher token usage thus does not guarantee higher completion across these models. The results motivate studying how agents use their available budget~\citep[cf.][]{simcot}, context, and tool interactions to satisfy the criteria for real-user revision requests.

\subsection{Further Analysis}
\label{sec:exp:further}

\subsubsection{Fine-Grained Capability Analysis}
\label{sec:exp:analysis}

Figure~\ref{fig:sec5_task_breakdown} summarizes performance across modalities and capabilities. On \trackone{}, mean strict completion is highest for document-grounded tasks ($65.8$), followed by the Image group ($64.2$), video ($63.4$), and audio ($59.0$). Audio-grounded revisions are therefore comparatively challenging within the benchmark's task mix.

Across capability groups (Figure~\ref{fig:sec5_task_breakdown}b), Content has the highest mean strict completion ($77.5$). Audio \& time ($59.5$), Style \& layout ($59.2$), Interaction ($59.5$), and Adaptation ($51.6$) all score below $60.0$. These results point to difficulties with temporal alignment, fine-grained layout, responsive adaptation, and multi-step interaction. Group sizes and task content differ, so these scores summarize the evaluated models' performance on the benchmark's task mix.

\begin{figure}[t]
\centering
\includegraphics[width=\textwidth]{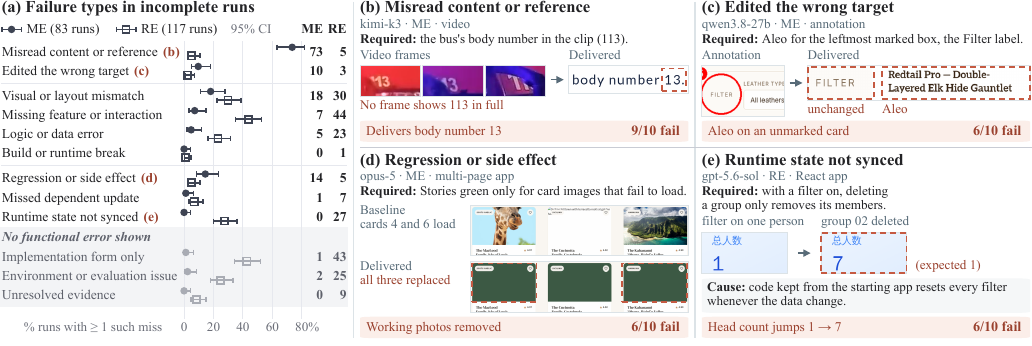}
\caption{\textbf{Failure types and recurring failure modes.} (a) Percentage of 200 randomly sampled incomplete runs (83 from \trackone{} and 117 from \tracktwo{}) that exhibit each failure type at least once. A run may exhibit multiple failure types. Grey rows represent failures that were not assigned an agent-error label. (b--e) Examples of final model outputs illustrating recurring failure modes. Each example illustrates a criterion failed by at least six of the original ten models (\trackone{}: b--d; \tracktwo{}: e). The title of each example indicates its failure type, corresponding to those in (a). The accompanying strips show the specific failure and the number of models that failed the criterion. See Appendix~\ref*{sec:app:failure_traj} for details.}
\label{fig:failure_types}\label{fig:failure_modes}
\end{figure}

\subsubsection{Completion Gap Analysis}
\label{sec:exp:gap}

Figure~\ref{fig:completion_gap}a shows the distribution of missed criteria among incomplete runs. On \trackone{}, $25.4\%$ of incomplete runs miss just one criterion, compared with $16.0\%$ on \tracktwo{}. Five or more misses occur in $34.4\%$ of incomplete \tracktwo{} runs, versus $13.1\%$ on \trackone{}. Repository edits therefore exhibit more failures with several missed criteria, although their rubrics are also longer. Allowing one missed criterion would raise the full-pass rate to $42.3\%$ on \trackone{} and $23.9\%$ on \tracktwo{}, illustrating how a near-complete rubric result can still fail full-task acceptance. Appendix~Table~\ref*{tab:completion_gap} reports the underlying counts.

Figure~\ref{fig:completion_gap}b shows that strict completion and full-pass rates do not move together. On \trackone{}, strict completion peaks at $69.4$ in the seven-to-eight-criterion group, while full-pass rates range from $32.8\%$ for two to four criteria to $10.0\%$ for nine to eleven. On \tracktwo{}, strict completion varies more modestly, and full-pass rates fall from $12.5\%$ in the smallest rubric group to $5.6\%$ in the largest. These comparisons involve different tasks; higher criterion completion need not imply a greater chance of passing the entire rubric.

In the archived cohort, tasks with five to eight criteria provide a comparison at matched rubric size ($66$ \trackone{} and $90$ \tracktwo{} tasks). Mean strict completion is similar ($60.4$ vs.\ $59.5$), but full-pass rates are $20.6\%$ and $10.4\%$, respectively. Full-pass rates differ by $10.2$ percentage points. Every model has a lower full-pass rate on the matched \tracktwo{} tasks. The gap therefore persists within this rubric-size band, although the tracks still differ in their requests, starting applications, and types of acceptance criteria (Appendix~\ref*{sec:app:gap}).

\subsubsection{Evaluation Reliability and Qualitative Analysis}
\label{sec:exp:reliability}
\label{sec:exp:failures}

\begin{table}[!htbp]
\centering
\caption{\textbf{Human--evaluator agreement.} Exact agreement and
linear-weighted Cohen's $\kappa$ between Opus 4.8 and
frontend-expert-reviewed reference labels on $100$ tasks:
$50$ \trackone{} tasks and $50$ \tracktwo{} tasks.
Both stages use the same outputs and reference.}
\label{tab:judge_proxy_agreement}
\footnotesize
\setlength{\tabcolsep}{2pt}
\renewcommand{\arraystretch}{1.12}
\begin{tabular*}{\linewidth}{@{\extracolsep{\fill}}lrrrrrr@{}}
\toprule
\multirow{2}{*}{\cdhead{Evaluator}}
& \multicolumn{3}{c}{\begin{tabular}[c]{@{}c@{}}\cdhead{Human--evaluator}\\\cdhead{agreement (\%) $\uparrow$}\end{tabular}}
& \multicolumn{3}{c}{\begin{tabular}[c]{@{}c@{}}\cdhead{Human vs. evaluator}\\\cdhead{Weighted Cohen $\kappa$ $\uparrow$}\end{tabular}} \\
\cmidrule(lr){2-4}\cmidrule(l){5-7}
 & \cdhead{Overall} & \cdhead{\trackone{}} & \cdhead{\tracktwo{}} & \cdhead{Overall} & \cdhead{\trackone{}} & \cdhead{\tracktwo{}} \\
\midrule
Judge & 64.69 & 69.93 & 61.01 & 0.518 & 0.611 & 0.454 \\
Judge-then-Refute & 66.04 & 71.90 & 61.93 & 0.538 & 0.641 & 0.468 \\
\bottomrule
\end{tabular*}
\end{table}

To evaluate the effectiveness of refutation, we conduct an additional evaluation on $100$ tasks, including $50$ from \trackone{} and $50$ from \tracktwo{}. We obtain initial (Judge) and post-refutation (Judge-then-Refute) verdicts for the same task outputs. We also ask frontend experts to independently review the same task outputs and provide reference judgments.

Table~\ref{tab:judge_proxy_agreement} compares evaluator judgments with the reference labels using exact agreement and linear-weighted Cohen's $\kappa$. Judge-then-Refute improves both measures on each track. Overall, exact agreement rises from $64.69\%$ to $66.04\%$, while weighted $\kappa$ rises from $0.518$ to $0.538$. Refutation thus yields a modest improvement on this sample, while substantial disagreement with the reference labels remains.

Figure~\ref{fig:failure_types}a summarizes an exploratory analysis of $713$ misses in $200$ randomly sampled current incomplete runs. LLM coding agents assigned one or more of twelve failure types to each miss, primarily from judge evidence; a run carries the union of its misses' types. The labels were not reviewed by humans (Appendix~\ref*{sec:app:failure_types}). These descriptive labels may reflect evaluator or annotator error and do not establish independently verified causes of failure.

The two tracks exhibit distinct failure patterns. In \trackone{}, $73\%$ of runs involve misreading content or references. In \tracktwo{}, the most common failure types are missing features or interactions ($44\%$), visual mismatches ($30\%$), unsynchronized runtime state ($27\%$), and logic errors ($23\%$). Regressions or missed dependent updates occur in $13.2\%$ of incomplete runs across both tracks (95\% interval: $[9.4,18.9]$). Including runtime-state errors increases this proportion to $29.8\%$ ($[24.0,36.6]$).

Figure~\ref{fig:failure_modes}b--e further illustrates four recurring failure modes through representative examples. Each example involves a criterion that at least six of the original ten models failed. We examined available final workspaces and solver trajectories for cases with clear requests and criteria. These examples show how seemingly reasonable editing decisions can produce an incorrect final application.

\noindent (b)~\textbf{Misread content or reference.} No frame of a parade video shows the bus number, 113, in full. Nine models fail this criterion, seven by misreading a partially obscured number (five write ``13''). The successful model recognizes that the number may be partially hidden, widens its crop, and searches other frames to obtain additional visual evidence.

\noindent (c)~\textbf{Edited the wrong target.} The request asks for a font change to the ``leftmost box'', marked by a circle around the filter label. Six models apply the requested change to a product card instead, indicating that they identify the requested modification but fail to locate its intended target.

\noindent (d)~\textbf{Regression or side effect.} The request is to replace only the card images that fail to load. Six models replace all card images, including two that load correctly. All four successful models first identify which image URLs fail, then replace those images while preserving the two that already load correctly in the starting application.

\noindent (e)~\textbf{Runtime state not synced.} The \tracktwo{} request asks for per-group deletion and requires all views to follow the active filter. In the starting app, all filters are reset whenever the data change. Six models retain this behavior, so deleting a group while a filter is active clears the filter and causes the head count to increase, for example, from 1 to 7. This case shows that implementing deletion alone does not ensure consistent filtering and counts in the final application.

These cases show why complete delivery requires grounding the requested change, checking affected dependencies, and verifying the final application state. Detailed trajectories and additional examples appear in Appendix~\ref*{sec:app:failure_traj}.

\section{Conclusion}
\label{sec:conclusion}

Co-Design Bench evaluates $273$ real-user application revisions through localized multimodal edits and coordinated repository changes, with runnable starting applications and $2{,}331$ hidden acceptance criteria. Under a shared judge-then-refute protocol, the strongest of ten endpoints reaches mean strict completion of $69.9$ but fully passes only $26.4\%$ of tasks. Our analyses reveal challenges in request grounding, change propagation, and final-state verification, motivating evaluation of both criterion-level progress and full-task acceptance for revisions requested by real users.

\bibliography{references}

\appendix
\raggedbottom
\widowpenalty=10000
\renewcommand{\topfraction}{0.9}
\renewcommand{\textfraction}{0.08}
\renewcommand{\floatpagefraction}{0.75}
\renewcommand{\dbltopfraction}{0.9}
\renewcommand{\dblfloatpagefraction}{0.75}
\setcounter{topnumber}{2}
\setcounter{dbltopnumber}{2}
\newcommand{\cdwidepage}[1]{%
  \clearpage\begingroup\captionsetup{skip=4pt,hypcap=false}%
  #1\par\endgroup\vspace{8pt}}
\makeatletter
\renewcommand\paragraph{\@startsection{paragraph}{4}{\z@}{6pt plus 1pt minus 1pt}{-1em}{\normalfont\normalsize\bfseries}}
\newcommand{\cdneedspace}[1]{\par\begingroup\@tempskipa=#1\relax
  \ifdim\dimexpr\pagegoal-\pagetotal\relax<\@tempskipa\newpage\fi\endgroup}
\makeatother

\cdwidepage{\centering
{\Large\textbf{Appendix}\par}\vspace{10pt}
\begin{minipage}{\linewidth}\raggedright
\subsection*{Guide to the Appendix}
\noindent\auditfix{\textbf{Evaluation reliability.} Judge sensitivity and repeatability (Sec.~\ref{sec:app:judgeswap}), harness sensitivity (Sec.~\ref{sec:app:cross_harness}), and human agreement and scoring procedures (Sec.~\ref{sec:app:eval_implementation}).}\par
\noindent\auditfix{\textbf{Performance and failures.} Modality, edit breadth, completion gaps, and refutation (Sec.~\ref{sec:app:detail}); failure types and solver trajectories (Sec.~\ref{sec:app:failure_traj}).}\par
\noindent\auditfix{\textbf{Tasks and user relevance.} Capability taxonomy (Sec.~\ref{sec:app:taxonomy}), task examples (Sec.~\ref{sec:app:cases}), dataset statistics (Sec.~\ref{sec:app:dataset_results}), and the questionnaire (Sec.~\ref{sec:app:questionnaire}).}\par
\noindent\auditfix{\textbf{Reproduction details.} Execution limits and resource accounting (Sec.~\ref{sec:app:implementation}), tool use (Sec.~\ref{sec:app:execution}), and data curation (Sec.~\ref{sec:app:curation}).}
\end{minipage}\par\vspace{8pt}
\centering
\includegraphics[width=0.92\linewidth]{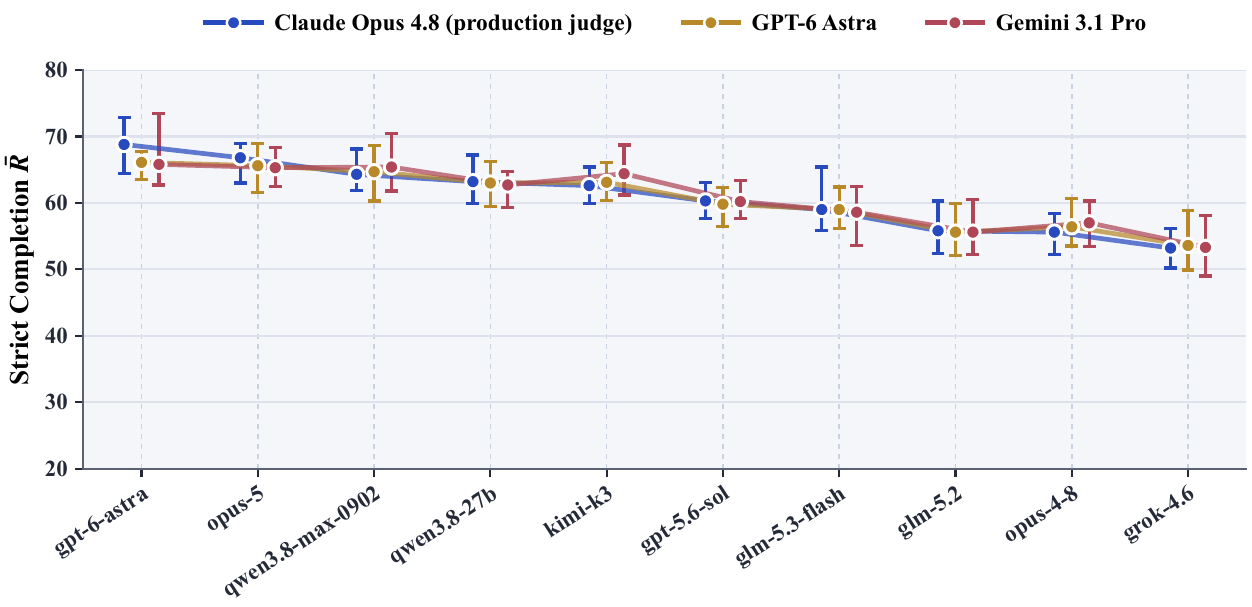}
\captionof{figure}{\textbf{Judge sensitivity on the fixed 32-task sample.} Strict Completion ($0$--$100$) across three judges; error bars indicate 95\% intervals. Models are ordered by the production judge's score.}
\label{fig:judge_sensitivity}}

\section{Judge Sensitivity and Stability}
\label{sec:app:judgeswap}

\paragraph{\auditfix{Judge comparison.}}
\auditfix{The three judges produce similar model-level score patterns on the fixed $32$-task sample (Figure~\ref{fig:judge_sensitivity}), suggesting that the broad comparison is not specific to a single evaluator.}

\paragraph{\auditfix{Repeatability.}}
\auditfix{We run the production judge three times on the same sample with identical inputs. Pairwise Pearson correlations between output-level Strict Completion scores range from $0.92$ to $0.96$, indicating a reproducible overall score pattern.}

\auditfix{The two checks vary different parts of the evaluation. Judge comparison changes the evaluator while holding the completed applications fixed; repeatability retains the production judge and repeats the evaluation on identical inputs. Together, they examine variation across evaluators and across repeated evaluations.}

\paragraph{\auditfix{Experimental setup.}}
\auditfix{The production evaluation pipeline uses \texttt{claude-opus-4-8} as both judge and refuter (Section~\ref*{sec:exp:setup}).} To examine sensitivity to judge choice, we re-score the final outputs of all ten models on a fixed $32$-task sample using the production judge, GPT-6~Astra~\citep{gpt6astra}, and Gemini~3.1~Pro~\citep{gemini31pro}. The sample contains $12$ \trackone{} and $20$ \tracktwo{} tasks, drawn by stratified random sampling over input modality and edit breadth, respectively. Each judge uses the same judging configuration and receives identical requests, hidden rubrics, baselines, and final workspaces. \auditfix{The comparison therefore reuses the same solver outputs across judges. Figure~\ref{fig:judge_sensitivity} summarizes model-level scores, whereas the repeatability correlations compare output-level scores across repeated evaluations.}


\cdwidepage{\section{Additional Results}
\label{sec:app:detail}

\subsection{Modality and Edit Scope}
\label{sec:app:strata}
\centering
\includegraphics[width=0.92\linewidth]{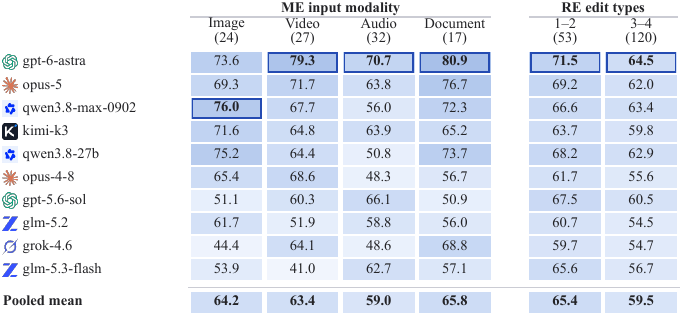}
\captionof{figure}{\textbf{Strict completion ($0$--$100$) by \trackone{} \auditfix{input modality} and \tracktwo{} edit breadth for the ten endpoints in Table~\ref*{tab:leaderboards}.} Rows follow current \trackone{} mean strict completion; outlines mark column maxima, and parentheses give task counts. \auditfix{The pooled modality means match Table~\ref{tab:t1_modality} and Figure~\ref*{fig:sec5_task_breakdown}a;} every endpoint scores lower on tasks with three or four edit types.}
\label{fig:modality_by_model}
\vspace{10pt}
\begingroup
\centering
\begin{minipage}[t]{0.62\linewidth}
\vspace{0pt}
\captionof{table}{\textbf{\trackone{} completion by \auditfix{input modality} across ten endpoints.}
Strict and Graded average task scores ($0$--$100$); verdict shares pool criterion judgments.
Full and Zero are task rates (\%) with $R_t=1$ and $R_t=0$.}
\label{tab:t1_modality}
\begingroup
\footnotesize
\centering
\setlength{\tabcolsep}{1.8pt}
\renewcommand{\arraystretch}{1.10}
\begin{tabular*}{\linewidth}{@{\extracolsep{\fill}}l V rr V rrr V r@{\extracolsep{0pt}\hspace{3pt}}r@{}}
\toprule
\multirow{2}{*}{\cdhead{Modality}} & \multicolumn{2}{cV}{\cdhead{Completion}} & \multicolumn{3}{cV}{\cdhead{Verdicts (\%)}} & \multicolumn{2}{c}{\makebox[0pt]{\cdhead{Outcomes}}} \\
\cmidrule(lr){2-3}\cmidrule(lr){4-6}\cmidrule(lr){7-8}
 & Strict & Graded & Pass & Partial & Fail & Full & Zero \\
\midrule
Image & 64.2 & 69.4 & 65.7 & 12.0 & 22.3 & 34.2 & 12.1 \\
Video & 63.4 & 70.0 & 63.9 & 13.0 & 23.1 & 19.6 & 1.5 \\
Audio & 59.0 & 65.3 & 59.6 & 12.4 & 28.0 & 20.9 & 0.3 \\
Document & 65.8 & 68.7 & 67.2 & 5.8 & 27.0 & 19.4 & 7.1 \\
\bottomrule
\end{tabular*}
\par\endgroup

\end{minipage}\hfill%
\begin{minipage}[t]{0.36\linewidth}
\vspace{0pt}
\captionof{table}{\textbf{\tracktwo{} completion by edit breadth across ten endpoints.}
Strict is mean task completion ($0$--$100$); Full and Zero are task rates (\%).}
\label{tab:edit_breadth}
\begingroup
\footnotesize
\centering
\setlength{\tabcolsep}{2pt}
\renewcommand{\arraystretch}{1.12}
\begin{tabular*}{\linewidth}{@{\extracolsep{\fill}}lrrr@{}}
\toprule
\cdhead{Edit types} & \cdhead{Strict} & \cdhead{Full} & \cdhead{Zero} \\
\midrule
1 & 60.8 & 10.0 & 0.0 \\
2 & 66.1 & 8.5 & 1.5 \\
3 & 59.6 & 8.7 & 2.5 \\
4 & 59.3 & 10.7 & 7.8 \\
\midrule
1--2 & 65.4 & 8.7 & 1.3 \\
3--4 & 59.5 & 9.8 & 5.5 \\
\bottomrule
\end{tabular*}
\par\endgroup

\end{minipage}
\par\endgroup
}

\paragraph{\auditfix{Results.}}
\auditfix{Document tasks have the highest pooled Strict Completion, video tasks the highest Graded Completion, and image tasks the highest Full-Pass Rate.} In \tracktwo{}, tasks spanning three or four edit types have lower Strict Completion and more zero-pass outcomes than those spanning one or two types, with similar full-pass rates; all ten endpoints score lower on the broader group, by $3.2$ to $8.9$ points of strict completion (Figure~\ref{fig:modality_by_model}). These groups differ in their requests and contexts as well as their labels; the comparisons describe the benchmark rather than estimate controlled effects of modality or edit breadth. \auditfix{The modality ranking changes with the completion metric: a group with more fully satisfied requests need not have the highest average criterion completion. Reporting Strict, Graded, and Full together makes this distinction visible.}

\paragraph{\auditfix{Groups and reporting.}}
Tables~\ref{tab:t1_modality} and~\ref{tab:edit_breadth} pool results across the ten endpoints by \trackone{} \auditfix{input modality} and \tracktwo{} edit breadth. \auditfix{Image includes screenshot-only tasks, consistent with the main Figure~\ref*{fig:sec5_task_breakdown}a.} Figure~\ref{fig:modality_by_model} gives each endpoint's strict completion by modality and edit breadth. Table~\ref{tab:strata} cross-tabulates \auditfix{input modality} and starting-artifact structure to clarify how these task attributes overlap.

\auditfix{The columns in Table~\ref{tab:t1_modality} use different units of evaluation. Strict and Graded average task-level scores, while verdict percentages pool individual criterion judgments. A task contributes one score to the completion average but can contribute several judgments to the verdict distribution. Full and Zero give the shares of tasks passing every criterion and passing none, respectively.}


\FloatBarrier
\subsection{Completion-Gap and Refutation Analysis}
\label{sec:app:gap}

\auditfix{We report the statistics underlying main Figure~\ref*{fig:completion_gap}: $2{,}730$ runs from ten models on $273$ tasks. Table~\ref{tab:completion_gap} gives miss counts and completion rates by rubric size.}

\begin{table}[!htbp]
\centering
\caption{\textbf{Misses and rubric size for main Figure~\ref*{fig:completion_gap}.} (a) Distribution of missed criteria among incomplete runs (\%); Runs gives their count, and $\le$1 miss gives the percentage of all runs missing at most one criterion. (b) Strict Completion ($0$--$100$) and Full-Pass Rate (\%) by rubric size.}
\label{tab:completion_gap}
\small
\setlength{\tabcolsep}{3.2pt}
\renewcommand{\arraystretch}{1.08}
\begin{tabular}{@{}l rrrrr r r@{}}
\toprule
\multicolumn{8}{@{}l}{\textbf{(a)} Misses per incomplete run} \\
\cdhead{Track} & \cdhead{1} & \cdhead{2} & \cdhead{3} & \cdhead{4} & \cdhead{5+} & \cdhead{Runs} & \cdhead{$\le$1 miss} \\
\midrule
\trackone{} & 25.4 & 27.2 & 20.2 & 14.2 & 13.1 & 773 & 42.3 \\
\tracktwo{} & 16.0 & 19.2 & 18.5 & 11.9 & 34.4 & 1{,}566 & 23.9 \\
\bottomrule
\end{tabular}
\par\vspace{5pt}
\begin{tabular}{@{}l rrr rrr@{}}
\multicolumn{7}{@{}l}{\textbf{(b)} Completion by criteria per task} \\
\toprule
 & \multicolumn{3}{c}{\cdhead{\trackone{}}} & \multicolumn{3}{c}{\cdhead{\tracktwo{}}} \\
\cmidrule(lr){2-4}\cmidrule(l){5-7}
\cdhead{Criteria} & \cdhead{Tasks} & \cdhead{Strict} & \cdhead{Full} & \cdhead{Tasks} & \cdhead{Strict} & \cdhead{Full} \\
\midrule
2--4 & 25 & 63.0 & 32.8 & -- & -- & -- \\
5--6 & 52 & 58.0 & 20.0 & 40 & 57.2 & 12.5 \\
7--8 & 14 & 69.4 & 22.9 & 50 & 61.2 & 8.8 \\
9--11 & 9 & 67.8 & 10.0 & 32 & 60.3 & 9.4 \\
12--15 & -- & -- & -- & 33 & 64.7 & 9.1 \\
16+ & -- & -- & -- & 18 & 63.5 & 5.6 \\
\midrule
5--8 \auditfix{(band)} & 66 & 60.4 & 20.6 & 90 & 59.5 & 10.4 \\
\bottomrule
\end{tabular}
\end{table}

\paragraph{Near misses and rubric size.}
\auditfix{Allowing one miss per run would raise the pooled full-pass rate from $22.7\%$ to $42.3\%$ on \trackone{}, from $9.5\%$ to $23.9\%$ on \tracktwo{}, and from $14.3\%$ to $30.7\%$ overall. On average, incomplete runs miss $49.5\%$ of their rubric on \trackone{} and $43.1\%$ on \tracktwo{}; the larger miss counts on \tracktwo{} also reflect its longer rubrics.}

\paragraph{\auditfix{Reading the miss distribution.}}
\auditfix{The miss-count bins and the $\leq 1$-miss column use different denominators. The five bins describe only incomplete runs: the one-miss entry is the share of these runs that fail exactly one criterion. The $\leq 1$-miss column includes all runs, combining those one-miss cases with runs that already pass the full rubric. It describes the coverage obtained by allowing one unfinished criterion. Because the tracks differ in rubric length, the miss-count distribution and size-stratified scores answer complementary questions: how much remains unfinished in a run, and whether completion differences persist among tasks with similar numbers of criteria.} \auditfix{In panel (b), the five-to-eight row combines the five-to-six and seven-to-eight groups; it is not an additional disjoint group. Its rates pool the corresponding runs before taking the track-level averages.}

\paragraph{\auditfix{Shared rubric-size band.}}
\auditfix{Within the five-to-eight-criterion band, strict completion is similar on \trackone{} and \tracktwo{} ($60.4$ and $59.5$), but full-pass rates are $20.6\%$ and $10.4\%$. The band contains $66$ \trackone{} and $90$ \tracktwo{} tasks. The difference (\tracktwo{} minus \trackone{}) is $-10.2$ percentage points, with a 95\% interval of $[-16.1,-4.0]$ from $10{,}000$ bootstrap resamples of tasks within each track (seed $20261110$). Each of the ten models has a lower full-pass rate on \tracktwo{} within this band. The difference also appears in the five-to-six band: strict completion is $58.0$ and $57.2$, while full-pass rates are $20.0\%$ and $12.5\%$. Exact rubric sizes and task content still differ between tracks within each band.}

\auditfix{Strict Completion measures the fraction of criteria satisfied, whereas Full-Pass Rate requires every criterion to be satisfied in the same run. Similar average criterion completion can therefore coexist with different rates of fully completed tasks, as the shared-band comparison illustrates.}

\paragraph{\auditfix{Refutation and human agreement.}}
\auditfix{On the separate $100$-task human-reference evaluation ($50$ tasks per track), refutation raises exact agreement from $64.69\%$ to $66.04\%$ and weighted Cohen's $\kappa$ from $0.518$ to $0.538$ (main Table~\ref*{tab:judge_proxy_agreement}). Section~\ref{sec:app:judge_proxy} describes the comparison procedure.}

\paragraph{\auditfix{Counting and aggregation.}}
\auditfix{A \emph{miss} is a criterion whose final verdict is not \textsc{Pass}; an \emph{incomplete} run does not pass the full rubric. We pool all ten models' runs, giving each task equal weight within its track. Rubric-size groups use the modal number of scored criteria per task; Strict Completion retains each run's recorded criterion count as its denominator. Runs without criterion records retain their task-level outcomes: three zero-score runs count all modal criteria as missed; nine full passes count none.}

\FloatBarrier
\section{Capability Taxonomy and\texorpdfstring{\\}{ }Multi-Label Coverage}
\label{sec:app:taxonomy}

\auditfix{Most tasks require multiple capabilities: $265$ of $273$ tasks ($97.1\%$) have multiple labels, and $174$ ($63.7\%$) span more than one high-level dimension (Figure~\ref{fig:task_taxonomy}). The taxonomy contains three dimensions, nine categories, and thirty fine-grained capabilities. One primary label identifies each task\textquotesingle s main delivery objective; additional labels capture its other requirements, yielding $1{,}172$ task--capability associations. Primary labels partition the task pool, while multi-label coverage records overlapping requirements.}

\auditfix{The Primary column assigns each task to its main delivery objective. The Any column includes every assigned capability, including the primary label, so one task can appear in several rows. These columns distinguish how often a capability defines the task from how often it participates in completing the request.}

\cdwidepage{\centering
\includegraphics[width=0.78\linewidth]{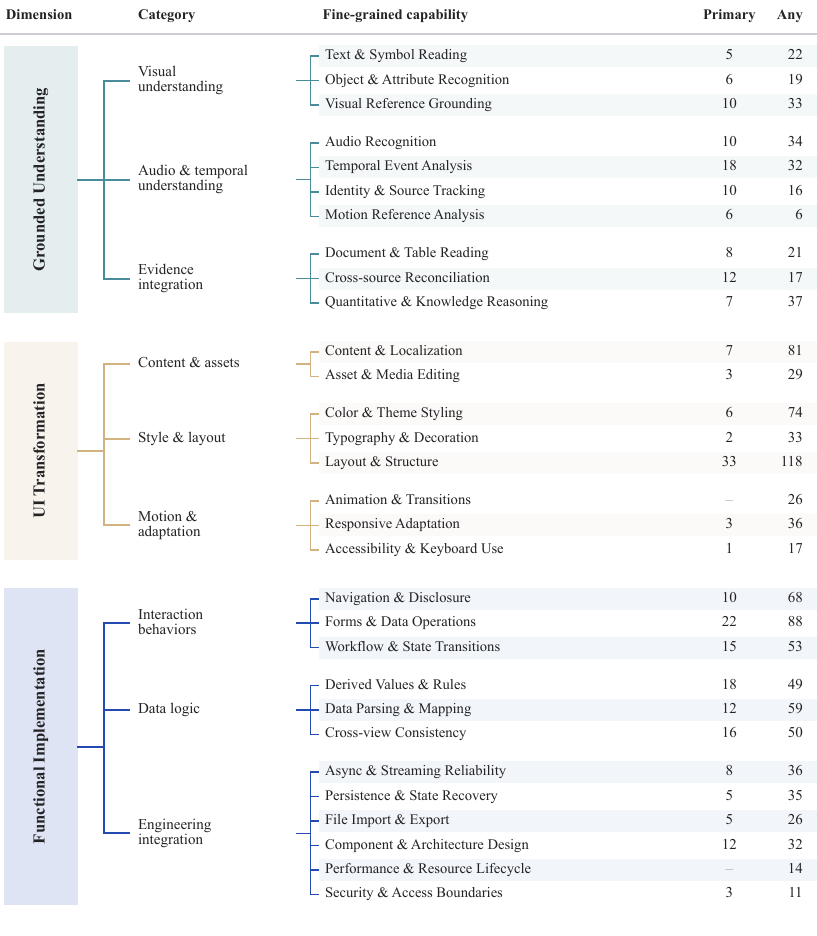}
\captionof{figure}{\textbf{Capability hierarchy and benchmark coverage ($N=273$).} Primary assignments partition tasks by main delivery goal; Any-label coverage includes auxiliary requirements, allowing overlap across categories. All thirty fine-grained capabilities are represented across the task suite.}
\label{fig:task_taxonomy}}

\section{Implementation and Resource Accounting}
\label{sec:app:implementation}
\label{sec:app:harness}

\paragraph{Execution constraints.}
Each trajectory has a three-hour wall-clock limit and a $120$-turn cap. A wall-clock timeout terminates execution and is recorded as an infrastructure failure. Reaching the turn cap instead triggers a build, render, and evaluation of the final workspace state.

\paragraph{Telemetry and token accounting.}
Solver token traffic comes from request-level API usage and includes retries and all solver attempts; evaluator traffic is excluded.
Turns count unique assistant messages, including subagents, in the final solver trace.
\par\noindent Len.\ averages each single-attempt rollout's request output tokens per turn; Tokens averages total traffic across scored cases with recorded usage. The available scored solver trajectories contain $135{,}451$ tool calls.

\paragraph{Trace integrity and final-state evaluation.}
Tool invocations and returns are paired sequentially within the same assistant scope. This prevents identifiers reused by nested agents from merging unrelated events. Intermediate renders and edits remain available for inspection. Scores use the final candidate: earlier behavior earns credit only if it remains present in the final workspace.

\cdwidepage{\section{Representative Benchmark Cases}
\label{sec:app:cases}
\centering
  \includegraphics[width=\linewidth]{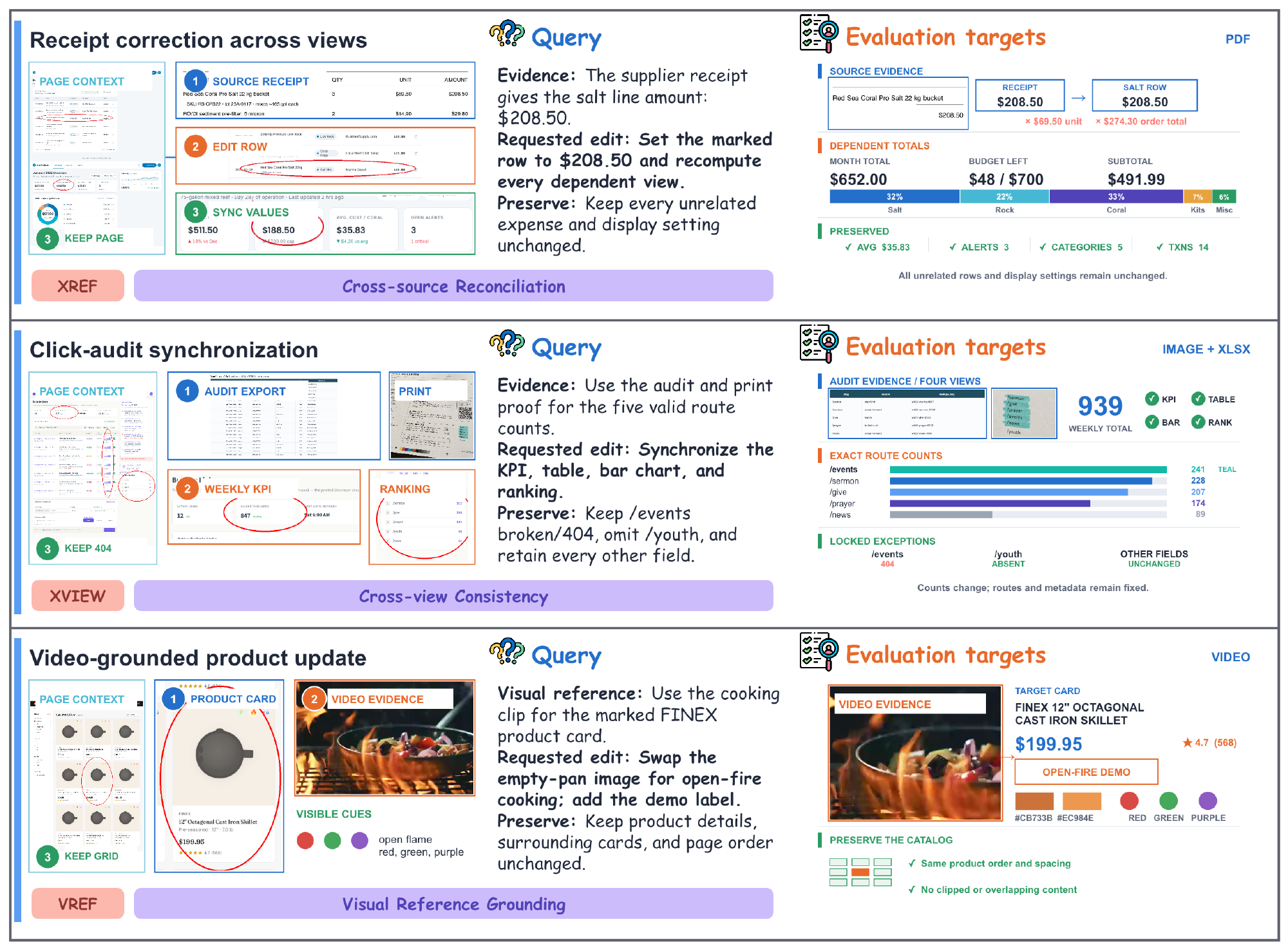}
  \captionof{figure}{\textbf{Representative \trackone{} cases: grounded local edits.}
  The examples cover cross-source reconciliation, cross-view consistency, and
  visual-reference grounding. Numbered markers organize the visible evidence;
  the right column visualizes three selected evaluator targets as value flows,
  compact charts, and state transitions. The complete rubric may contain additional criteria.}
  \label{fig:track1_cases}}

\auditfix{Figures~\ref{fig:track1_cases} and~\ref{fig:track2_cases} illustrate grounding, dependent updates, and preservation constraints. Each row connects the starting application and supporting evidence (left), the request (center), and three selected acceptance criteria (right). The examples illustrate the task interface; evaluation uses the full task pool and complete hidden rubrics.}

\paragraph{What the \trackone{} examples test.}
\auditfix{The receipt case links the supplier's source document to the marked salt row. The row must use the corresponding amount rather than a unit price or order total. Updating it must also change the month total, remaining budget, and category subtotal, while unrelated expenses and display settings stay fixed. The click-audit case extends this dependency check across a weekly KPI, table, bar chart, and ranking.}
\par\noindent
\auditfix{The audit export and print proof establish the valid route counts, which must agree across these views. The preservation boundary is explicit: \texttt{/events} retains its broken status, \texttt{/youth} stays omitted, and other fields remain unchanged. In the video-grounded case, the cooking clip supplies the visual reference for the marked FINEX product card. The replacement changes the empty-pan image to open-fire cooking and adds the demo label, while keeping the product name, price, and rating. Neighboring cards, product order, and spacing must remain intact, without clipped or overlapping content. Across these examples, the selected checks connect the supplied evidence to both the requested updates and the application content that those updates must preserve.}

\cdwidepage{\centering
  \includegraphics[width=\linewidth]{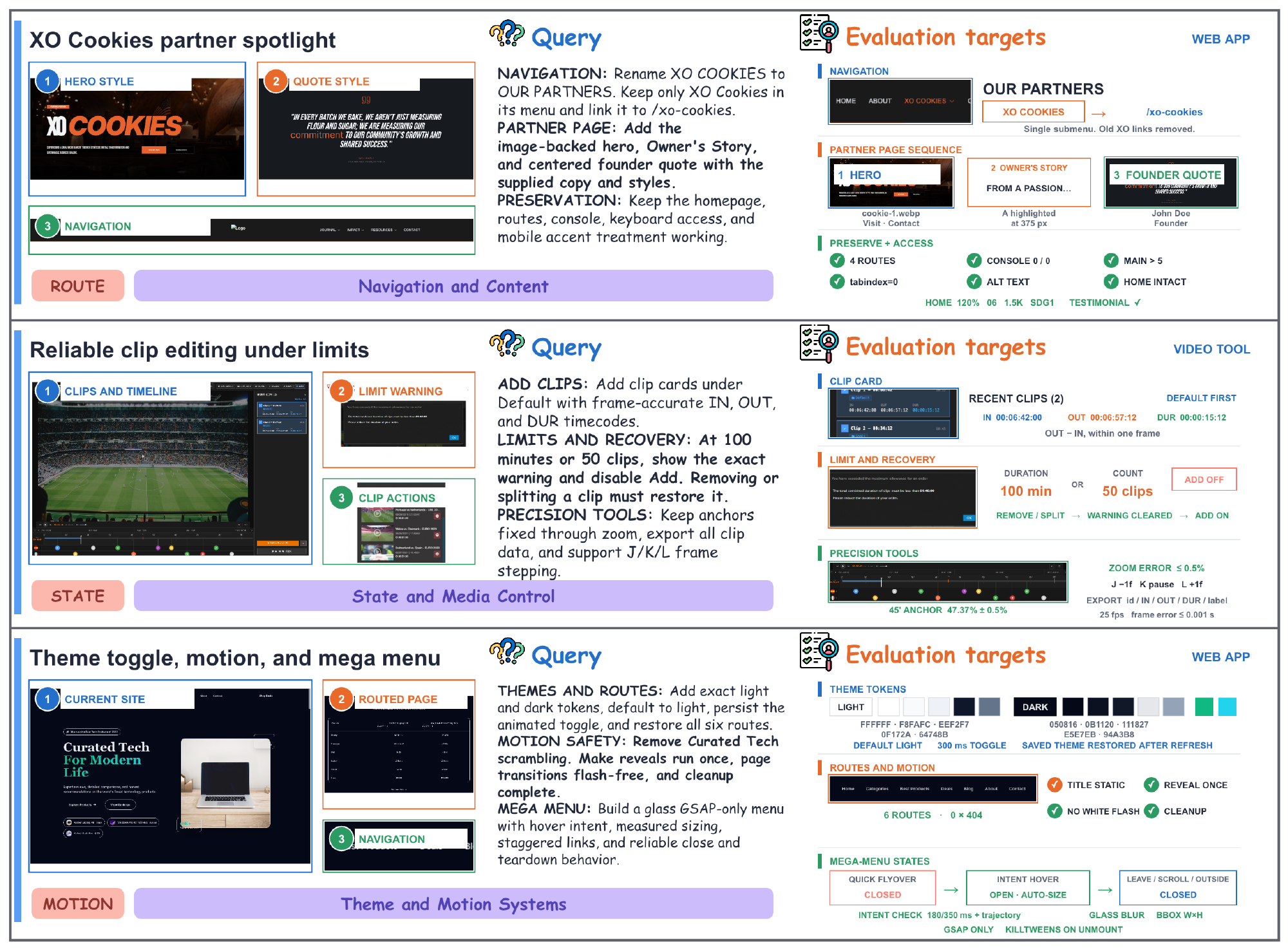}
  \captionof{figure}{\textbf{Representative \tracktwo{} cases: coordinated repository edits.}
  The examples span navigation and page construction, clip editing under
  resource limits, and theme-aware motion with a morphing mega menu. The
  selected targets use route flows, thresholds, token palettes, and interaction
  states to make the behavior and preservation constraints explicit.}
  \label{fig:track2_cases}}

\paragraph{What the \tracktwo{} examples test.}
\auditfix{The partner-page case uses the supplied hero and quote styles together with the requested navigation structure to define a coordinated change. The existing XO COOKIES entry becomes OUR PARTNERS, with a single XO Cookies submenu linking to \texttt{/xo-cookies}. That destination must present the image-backed hero, Owner's Story, and centered founder quote using the supplied copy and styling. These additions are constrained by the existing application: the homepage, other routes, keyboard access, and mobile accent treatment must continue to work. The requirements cross page boundaries: adding the destination alone would leave the navigation change incomplete, while changing the menu alone would leave its promised content unavailable. The clip-editor case shifts the focus from page structure to state. Clip cards expose IN, OUT, and duration values, so the displayed timing must remain consistent with the selected interval and the frame-accurate controls.}
\par\noindent
\auditfix{The editor's resource limits couple visible warnings to available actions: reaching the duration or clip-count limit disables Add, while removing or splitting a clip must support recovery. The task also requires stable timeline anchors through zoom, export of clip data, and keyboard frame stepping. In the theme case, the requested light and dark tokens must work with a persistent animated toggle and restored routes. Motion changes include a static title, reveals that run once, and transitions without a white flash. The requested GSAP mega menu adds another state sequence: a quick flyover stays closed, an intentional hover opens a measured panel, and leaving, scrolling, or clicking outside closes it. Cleanup and teardown complete that behavior. Together, these examples connect design and interaction requirements to changes that remain coherent across navigation, editing actions, and repeated interface transitions.}

\cdwidepage{\section{Dataset Composition and Category-Level Results}
\label{sec:app:dataset_results}
\subsection{Dataset Composition}
\label{sec:app:composition}}

\begingroup\captionsetup{skip=4pt,hypcap=false}
\begingroup
\centering
\captionof{table}{\textbf{Composition of the 273-task benchmark.} Percentages use each track\textquotesingle s task count; \tracktwo{} edit types overlap. Additional attachments refer to \trackone{}, captures to \tracktwo{}. Mean captures use tasks with captures.}
\label{tab:dataset_stats}
\small
\setlength{\tabcolsep}{3pt}
\renewcommand{\arraystretch}{1.20}
\begin{tabular}{@{}>{\raggedright\arraybackslash}p{\dimexpr\linewidth-124pt\relax} >{\centering\arraybackslash}p{56pt} >{\centering\arraybackslash}p{56pt}@{}}
\toprule
 & \textbf{\trackone{}} & \textbf{\tracktwo{}} \\
 & \emph{design edit} & \emph{whole-app\newline reflow} \\
\midrule
Tasks & $100$ & $173$ \\
Evaluation criteria & $553$ & $1{,}778$ \\
Per task:\newline mean / median / max & $5.5$ / $6$ / $11$ & $10.3$ / $8$ / $46$ \\
\midrule
\auditfix{With add. attachments / captures} & $80$ ($80\%$) & $142$ ($82\%$) \\
\auditfix{Mean add. attachments / captures} & $1.35$ & $5.1$ \\
\midrule
\multicolumn{3}{@{}l}{\emph{\trackone{}: input modality}} \\
\quad Image & $24$ ($24\%$) & -- \\
\quad Video & $27$ ($27\%$) & -- \\
\quad Audio & $32$ ($32\%$) & -- \\
\quad Document & $17$ ($17\%$) & -- \\
\midrule
\multicolumn{3}{@{}l}{\emph{\tracktwo{}: edit types}} \\
\quad Structure & -- & $151$ ($87\%$) \\
\quad Style & -- & $133$ ($77\%$) \\
\quad Interactive & -- & $131$ ($76\%$) \\
\quad Content & -- & $112$ ($65\%$) \\
\quad Multiple edit types & -- & $166$ ($96\%$) \\
\quad Interaction depth\newline\hspace*{1em}L3--L4 & -- & $101$ ($58\%$) \\
\midrule
Starting artifact & Single page $79$;\newline bundled app $21$ & React $171$\newline ($99\%$) \\
\bottomrule
\end{tabular}
\par\endgroup

\endgroup
\par\vspace{6pt}
\auditfix{\trackone{} modality labels describe the evidence used to interpret a requested edit, whereas \tracktwo{} edit types describe the changes required in the application. The \tracktwo{} labels overlap, so one task may combine structural, stylistic, interactive, and content changes. The prevalence of multiple edit types therefore captures the need to satisfy several kinds of requirement within the same task. Rubric size adds a separate description: \tracktwo{} has more criteria per task on average, and its mean exceeds its median. The counts describe how the requested work is divided into individually evaluated requirements, alongside the modality and edit-type labels.}
\begingroup\captionsetup{skip=4pt,hypcap=false}
\begingroup
\centering
\captionof{table}{\textbf{Starting artifact and \auditfix{input modality} on \trackone{}.} Shading shows task counts; dashes denote zero. Strict ($\bar R$, $0$--$100$) averages task scores across ten endpoints; grey scores represent single-task groups. The bundled subtotal includes both source-tree rows.}
\label{tab:strata}
\small
\setlength{\tabcolsep}{1.5pt}
\renewcommand{\arraystretch}{1.15}
\begin{tabular*}{\linewidth}{@{\extracolsep{\fill}}l cccc r r@{}}
\toprule
& \multicolumn{4}{c}{\cdhead{\auditfix{Input modality}}} & & \\
\cmidrule(lr){2-5}
\cdhead{Starting artifact} & Image & Video & Audio & Doc. & \cdhead{Total} & \cdhead{Strict}\cdup \\
\midrule
Single-file HTML & \cdhm{cdwqwen}{19}{6} & \cdhm{cdwqwen}{42}{25} & \cdhm{cdwqwen}{50}{32} & \cdhm{cdwqwen}{31}{16} & 79 & 62.8 \\
Bundled app & \cdhm{cdwqwen}{33}{18} & \cdna & \cdna & \cdhm{cdwqwen}{13}{1} & 19 & 63.4 \\
\textsc{React} source tree & \cdna & \cdhm{cdwqwen}{13}{1} & \cdna & \cdna & 1 & \textcolor{cdmuted}{44.0} \\
\textsc{Vue} source tree & \cdna & \cdhm{cdwqwen}{13}{1} & \cdna & \cdna & 1 & \textcolor{cdmuted}{51.1} \\
\addlinespace[1.5pt]
\emph{Bundled subtotal} & \cdhm{cdwqwen}{33}{18} & \cdhm{cdwqwen}{14}{2} & \cdna & \cdhm{cdwqwen}{13}{1} & 21 & 61.9 \\
\midrule
\emph{All tasks} & 24 & 27 & 32 & 17 & 100 & 62.6 \\
\bottomrule
\end{tabular*}
\par\endgroup

\endgroup
\vspace{6pt}

Table~\ref{tab:dataset_stats} reports task counts, criterion density, reference materials, and starting-artifact metadata. Among \tracktwo{} tasks with capture artifacts, the mean is $5.1$ captures per task. Table~\ref{tab:strata} shows that \trackone{} contains $79$ single-page HTML applications ($79\%$) and $21$ bundled multi-file application trees. In \tracktwo{}, $171$ tasks ($98.8\%$) use single-page React applications. A single-page application can still require coordinated edits across several repository files.

\subsection{Category-Level Completion Breakdown}

\auditfix{Content and Assets has the highest pooled Strict Completion ($77.5$), while Motion and Adaptation has the lowest ($51.6$). Figure~\ref{fig:capability_matrix} gives endpoint- and category-level results for the current evaluation grid in Table~\ref*{tab:leaderboards}; its pooled and full-pass columns appear in the main Figure~\ref*{fig:sec5_task_breakdown}b.} The Motion and Adaptation category contains only four tasks, so its scores summarize a particularly small group. Capability labels also align with the two tracks: $89$ of $92$ Grounded Understanding tasks belong to \trackone{}, while $123$ of $126$ Functional Implementation tasks belong to \tracktwo{}. Category differences therefore combine capability requirements with the editing contexts in which those requirements occur.

\auditfix{Within each category, the model columns show variation across endpoints, while the pooled score summarizes their task-level results. Full Pass asks whether all criteria are satisfied in the same task. A category may therefore combine relatively high mean completion with a much lower full-pass rate. Reading these columns together distinguishes progress on individual requirements from complete execution of the requested change.}

\clearpage
\begingroup
\centering
\captionsetup{skip=4pt,hypcap=false}
\includegraphics[width=0.90\linewidth]{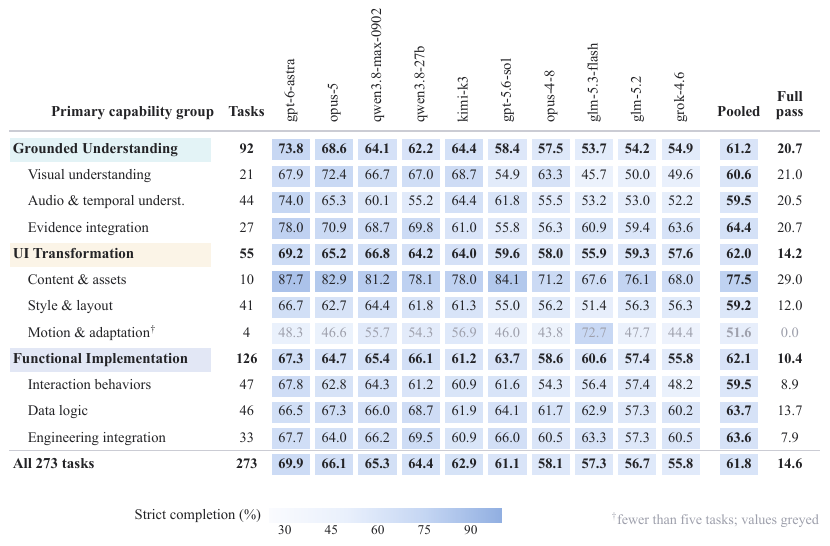}
\captionof{figure}{\textbf{Mean strict completion ($\bar R$) across primary capability categories and model endpoints (\%).} Bold rows denote top-level dimensions; sub-rows report constituent categories. Pooled columns report overall average scores across all evaluated pairs, while Full Pass ($\mathrm{F}$) denotes the share of completely resolved tasks. The low-sample Motion and Adaptation category ($\dagger$, $n=4$) is displayed in grey. Category names are longer forms of the short labels in Figures~\ref*{fig:teaser}a and~\ref*{fig:sec5_task_breakdown}b of the main paper.}
\label{fig:capability_matrix}
\vspace{10pt}
\centering
  \includegraphics[width=0.80\linewidth]{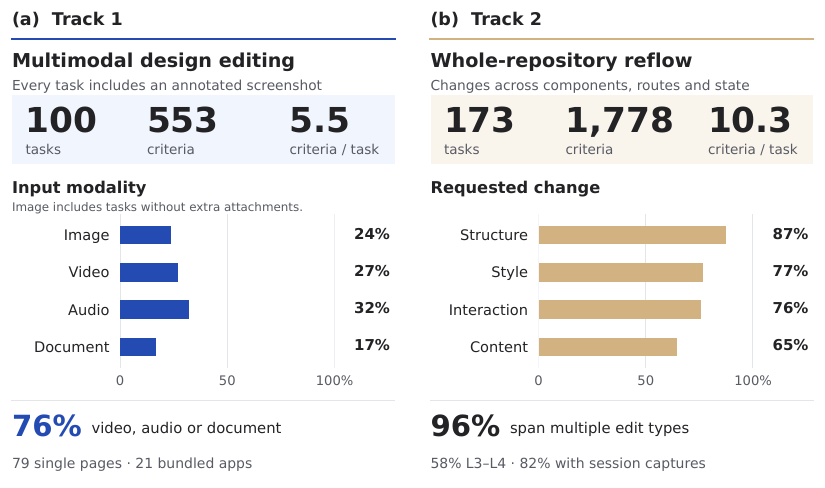}
  \captionof{figure}{\textbf{Two complementary tracks in Co-Design Bench.} Both originate in real user coding sessions. \trackone{} groups tasks by \auditfix{input modality}; \tracktwo{} change types overlap. Bars show shares within each track; criteria per task is the mean. Detailed statistics appear in Table~\ref{tab:dataset_stats}.}
  \label{fig:dataset_overview}
\par\endgroup

\cdwidepage{
\begingroup
\centering
\captionof{table}{\textbf{Identifiable media-processing Bash requests by modality.} \auditfix{The archived media audit covers $1{,}000$ final runs of ten endpoints on the released \trackone{} tasks.} Each cell gives the number of Bash requests and, in parentheses, the percentage of runs with at least one such request. A request counts once per named tool; columns can overlap. Installation, import/version checks, and script preparation alone are excluded; failed processing requests are included.}
\label{tab:media_tools}
\footnotesize
\setlength{\tabcolsep}{5pt}
\renewcommand{\arraystretch}{1.12}
\begin{tabular*}{\linewidth}{@{\extracolsep{\fill}}lrrrr@{}}
\toprule
\cdhead{Modality (tasks)} & \cdhead{Runs} & \cdhead{\texttt{ffmpeg}} & \cdhead{OpenCV} & \cdhead{\texttt{faster-whisper}} \\
\midrule
No attachment (20) & 200 & 1 (0.5\%) & 0 (0.0\%) & 0 (0.0\%) \\
Image (4) & 40 & 0 (0.0\%) & 0 (0.0\%) & 0 (0.0\%) \\
Video (27) & 270 & 257 (31.5\%) & 2{,}438 (68.5\%) & 20 (3.3\%) \\
Audio (32) & 320 & 160 (25.0\%) & 3 (0.3\%) & 65 (4.4\%) \\
Document (17) & 170 & 0 (0.0\%) & 96 (3.5\%) & 0 (0.0\%) \\
\midrule
All \trackone{} (100) & 1000 & 418 (16.6\%) & 2{,}537 (19.2\%) & 85 (2.3\%) \\
\bottomrule
\end{tabular*}
\par\endgroup
}

\section{Tool Use and Media Processing}
\label{sec:app:execution}

\begin{table}[!htbp]
\centering
\caption{\textbf{Tool-call shares by track (\%).} Share of \auditfix{all recorded tool calls for an endpoint on a track}: execution (Exec, \texttt{Bash}), inspection (Insp., \texttt{Read/Grep/Glob}), and modification (Mod., \texttt{Edit/Write}); auxiliary tools make up the remainder. Rows follow Table~\ref*{tab:leaderboards}.}
\label{tab:tool_shares}
\footnotesize
\setlength{\tabcolsep}{3.2pt}
\renewcommand{\arraystretch}{1.1}
\begin{tabular}{@{}l rrr V rrr@{}}
\toprule
\multirow{2}{*}{\cdhead{Model}} & \multicolumn{3}{cV}{\cdhead{\trackone{}}} & \multicolumn{3}{c}{\cdhead{\tracktwo{}}} \\
\cmidrule(lr){2-4}\cmidrule(l){5-7}
 & Exec & Insp. & Mod. & Exec & Insp. & Mod. \\
\midrule
\cdmi{\cdiconopenai}{gpt-6-astra} & 36.8 & 34.7 & 13.7 & 20.3 & 31.1 & 37.6 \\
\cdmi{\cdiconanthropic}{opus-5} & 55.8 & 35.8 & 7.8 & 44.4 & 19.2 & 34.5 \\
\cdmi{\cdiconqwen}{qwen3.8-max-0902} & 46.9 & 33.0 & 7.8 & 32.7 & 22.5 & 28.3 \\
\cdmi{\cdiconqwen}{qwen3.8-27b} & 45.8 & 35.7 & 6.6 & 32.8 & 28.0 & 22.3 \\
\cdmi{\cdiconkimi}{kimi-k3} & 47.9 & 40.0 & 7.9 & 34.3 & 27.4 & 30.7 \\
\cdmi{\cdiconopenai}{gpt-5.6-sol} & 34.1 & 35.7 & 9.5 & 20.8 & 36.8 & 19.6 \\
\cdmi{\cdiconanthropic}{opus-4-8} & 49.8 & 37.8 & 9.9 & 27.6 & 25.3 & 32.5 \\
\cdmi{\cdiconglm}{glm-5.3-flash} & 53.6 & 40.7 & 3.7 & 43.0 & 21.6 & 29.6 \\
\cdmi{\cdiconglm}{glm-5.2} & 74.6 & 13.1 & 4.9 & 38.7 & 24.2 & 18.2 \\
\cdmi{\cdiconxai}{grok-4.6} & 33.6 & 54.2 & 12.2 & 2.5 & 58.9 & 38.6 \\
\bottomrule
\end{tabular}
\end{table}

\paragraph{\auditfix{Cross-track tool activity.}}
\auditfix{\trackone{} runs use more image inspection, while \tracktwo{} runs devote a larger share of tool calls to modification. Pooled over the ten models, image-inspection calls average $14.61$ on \trackone{} and $2.79$ on \tracktwo{} (over $1{,}720$ traced \tracktwo{} runs). \texttt{Edit}/\texttt{Write} accounts for $7.4\%$ and $27.7\%$ of all calls, respectively (Table~\ref{tab:tool_shares}).}

\paragraph{\auditfix{Long trajectories.}}
\auditfix{On \trackone{}, video tasks take a median of $44.5$ turns, against $27$ for tasks without additional attachments. Audio and video tasks make up $82.8\%$ of each endpoint's longest quartile on average, compared with $34.4\%$ of the shortest quartile and $59\%$ of all \trackone{} tasks. To form the quartiles, we sort scored rollouts within each endpoint by turn count, break ties by task ID, and average endpoint shares equally.}

\paragraph{\auditfix{Counting conventions.}}
\auditfix{Image-inspection calls include attachment reads, extracted video frames, and browser screenshots; modification calls include \texttt{Edit}/\texttt{Write}. Tool statistics use available traces, while completion scores retain all $1{,}730$ \tracktwo{} model--task pairs. These summaries describe tool use rather than the effect of a tool family on success.}

\paragraph{Media processing.}
Table~\ref{tab:media_tools} reports named media-processing tools from the archived ten-endpoint \trackone{} media audit. The categories include direct \texttt{ffmpeg} calls with specified input streams, OpenCV image or video routines, and \texttt{faster-whisper} transcription. Generated helper scripts count when the trace records their subsequent execution in the shell.

\subsection{External Pretrained Models}
\label{sec:app:hf_models}

Table~\ref{tab:hf_models} gives examples from five solver endpoints that download or execute pretrained perception models from Hugging Face. The audit covers all ten endpoints, including \texttt{gpt-6-astra}. Observed outputs include AST audio-class scores, CLAP audio--text similarity scores, Whisper transcripts, and Qwen2-Audio descriptions. The selected rows illustrate available processing strategies; their counts are not prevalence estimates.

\begin{table}[!t]
\begingroup
\centering
\captionof{table}{\textbf{Hugging Face models in selected final trajectories.}
Rows group external models and observed outputs by solver.
AST is the Audio Spectrogram Transformer fine-tuned on AudioSet;
Whisper variants are invoked through \texttt{faster-whisper}.}
\label{tab:hf_models}
\footnotesize
\setlength{\tabcolsep}{5pt}
\renewcommand{\arraystretch}{1.12}
\begin{tabular}{@{}>{\raggedright\arraybackslash}p{0.23\linewidth}
                  >{\raggedright\arraybackslash}p{0.29\linewidth}
                  >{\raggedright\arraybackslash}p{\dimexpr0.48\linewidth-4\tabcolsep\relax}@{}}
\toprule
\cdhead{Solver} & \cdhead{External models} & \cdhead{Observed evidence} \\
\midrule
\texttt{gpt-6-astra} &
AST (AudioSet), CLAP\newline(HTSAT-unfused),\newline Whisper small,\newline Qwen2-Audio-7B-Instruct &
AudioSet class scores, audio--text similarity scores, timestamped transcripts,
and generated audio descriptions. \\
\addlinespace[3pt]
\texttt{opus-5} &
AST (AudioSet),\newline Whisper small &
Audio class scores (e.g., Frog 0.91 and Church bell 0.84) and timestamped Chinese transcripts. \\
\addlinespace[3pt]
\texttt{kimi-k3} &
AST and YAMNet\newline(ONNX weights) &
AST weights downloaded; YAMNet inference returns animal-class scores over time. \\
\addlinespace[3pt]
\texttt{glm-5.3-flash} &
CLIP ViT-B/32,\newline DINOv2-small,\newline OWLv2-base (ensemble) &
Video-frame scene scores, crop embedding similarities, and object detection boxes. \\
\addlinespace[3pt]
\texttt{glm-5.2} &
Qwen2-Audio-7B-Instruct,\newline AST (AudioSet),\newline Whisper base, large-v3 &
Generated audio descriptions and timestamps, AudioSet class scores,
and timestamped transcripts. \\
\bottomrule
\end{tabular}
\par\endgroup

\end{table}

Returned inference payloads distinguish successful model use from attempted execution. In one OWLv2 example, a post-processing exception occurs despite a zero shell exit code; a corrected call subsequently returns detection boxes. The traces thus document concrete uses of external perception models through local runtimes, including PyTorch, ONNX Runtime, and \texttt{faster-whisper}, without establishing how much these calls improve task completion.

\section{Cross-Harness Sensitivity}
\label{sec:app:cross_harness}

\paragraph{\auditfix{Results.}}
Tab.~\ref{tab:cross_harness} summarizes Strict Completion across the three harnesses. \texttt{gpt-6-astra} ranges from $63.1$ to $67.4$, a spread of $4.3$ points, while \texttt{kimi-k3} ranges from $59.5$ to $63.0$, a spread of $3.5$ points. \texttt{gpt-6-astra} scores higher under all three harnesses, by $2.2$, $7.9$, and $2.0$ points under Claude Code, Codex, and OpenCode, respectively, preserving the relative ordering of the two models. These results suggest that harness choice affects absolute scores, while the observed model ordering remains consistent on this $32$-task sample. We therefore report model and harness together and hold the harness fixed for the main comparison.

\begin{table}[!htbp]
\centering
\caption{\textbf{Cross-harness sensitivity on the fixed 32-task sample.}}
\label{tab:cross_harness}
\small
\setlength{\tabcolsep}{6pt}
\renewcommand{\arraystretch}{1.12}
\begin{tabular*}{0.95\linewidth}{@{\extracolsep{\fill}}l l c@{}}
\toprule
\cdhead{Model} & \cdhead{Harness} & \cdhead{Strict} \\
\midrule
\cdmodel{gpt-6-astra} & Claude Code & 65.2 \\
 & Codex & 67.4 \\
 & OpenCode & 63.1 \\
\addlinespace[2pt]
\cdmodel{kimi-k3} & Claude Code & 63.0 \\
 & Codex & 59.5 \\
 & OpenCode & 61.1 \\
\bottomrule
\end{tabular*}
\end{table}

\paragraph{\auditfix{Experimental setup.}}
The main comparison uses a shared Claude Code harness across models. To assess sensitivity to the harness, we evaluate \texttt{gpt-6-astra} and \texttt{kimi-k3} under Claude Code~\citep{claude_code} (version 2.1.152), Codex~\citep{codex}, and OpenCode~\citep{opencode}. We use the same fixed $32$-task subset as the judge-sensitivity study (Sec.~\ref{sec:app:judgeswap}), comprising $12$ \trackone{} and $20$ \tracktwo{} tasks. The subset was selected before re-scoring, without using scores, by stratified random sampling with proportional allocation over input modality (\trackone{}) and edit breadth (\tracktwo{}).

For each task, all three harnesses receive identical task text, attachments, and initial workspaces. All runs use low reasoning effort, a three-hour solve limit, and the same infrastructure-retry policy as the main setting (Section~\ref*{sec:exp:setup}). Each harness retains its native system prompt, tools, and context management. All outputs are scored by the production judge using the protocol of Section~\ref*{sec:eval}.

\FloatBarrier
\section{Data Curation Details}
\label{sec:app:curation}

\auditfix{Curation proceeds through environment recovery, request filtering, candidate ranking, and session consolidation before rubric authorship. The following paragraphs detail the pipeline summarized in Section~\ref*{sec:benchmark:source}.}

\paragraph{Source corpus and notation.}
The corpus contains $39{,}325$ query records from $26{,}314$ coding sessions, including $25{,}373$ frontend application sessions. Deduplication leaves $37{,}463$ candidates, each anchored to an edit turn, repository, available history, and attachments. In Figure~\ref*{fig:curation_pipeline}, $Q\mathbin{\cdot}I\mathbin{\cdot}B\mathbin{\cdot}C$ is shorthand: $Q=q$, the figure's $I=(X,K)$ combines attachments and context, and the manifest bundled with $B$ specifies environment $E$.

\paragraph{Environment recovery and validation.}
Starting applications are recovered from telemetry logs, repository archives, standard scaffolds, or static HTML bundles. Recovery restores code, assets, and available history while recording the original source hash. Validation checks code integrity, dependencies, builds, container launch, and headless Playwright rendering. Candidates requiring inaccessible private APIs, authenticated backends, or proprietary SDKs are excluded. Other faulty repositories undergo P0 scriptable batch repair, P1 agentic repair, or P2 structural pruning, followed by the same validation gate. A validated package contains the baseline, request context, attachments, container manifest, and validation logs.

\paragraph{Deduplication and filtering.}
Content hashes identify exact duplicates; near-duplicates with tokenized Jaccard similarity above $0.80$ are removed. Normalization decodes attachments and replaces local image paths with packaged asset references. An optional OCR and content-inspection module screens confidential references while retaining task specifications. Before ranking, deterministic filters F1--F6 reject environment-blocked turns, non-actionable chatter, unbounded requests, system trajectory artifacts, and non-editing interactions. Candidates must pass every applicable filter.

\paragraph{Candidate ranking.}
Surviving queries receive a heuristic score combining structural difficulty, verifiability, and historical failure signals:
\begin{equation}
  \begin{gathered}
  r(q)=d(q)+v(q)+f(q),\\
  d\in[0,5],\;v\in[0,3],\;f\in[0,2].
  \end{gathered}
  \label{eq:curation_rank}
\end{equation}

The structural difficulty score prioritizes coordination requirements: $d=1$ denotes localized cosmetic edits such as changing one CSS property; $d=2\text{--}3$ denotes dynamic client-side state changes or intra-component reflow; and $d=4\text{--}5$ denotes multi-component refactoring, cross-file data propagation, or complex edge cases. Verifiability depends on the required inspection: static source diffs ($v=1$), rendered visual DOM properties ($v=2$), or dynamic post-interaction state transitions ($v=3$); ambiguous or non-auditable targets receive $v=0$. Historical failure signals are scored after removing environment-blocked turns: a single-turn resolution receives $f=0$; two or three query reformulations receive $f=1$; and four or more repetitive follow-up attempts or an explicit rollback receive $f=2$.

\paragraph{Session consolidation and rubric authorship.}
The selector keeps at most three eligible queries ($r(q)>1$) per session. Their bounding message indices define a closed interval that retains intervening user turns in chronological order. Sessions with $\text{question\_total\_score}\ge10$ proceed to consolidation. For \tracktwo{}, experts refine the request, assign metadata, and author hidden criteria based on the selected user intent. Solver packages contain the task description and working copy; evaluator packages additionally retain the rubrics.

\section{Evaluation Implementation Details}
\label{sec:app:eval_implementation}

The evaluator receives the request $q$, hidden rubric $C$, baseline $B$, candidate application $W$, and execution logs. The procedures below implement the evidence requirements in Section~\ref*{sec:eval}.

\paragraph{Human--evaluator agreement.}
\label{sec:app:judge_proxy}
\auditfix{On $100$ saved tasks ($50$ per track), we compare Opus~4.8's initial and post-refutation verdicts against reference labels checked by frontend experts, using the same outputs and references for both stages. Refutation raises overall exact agreement from $64.69\%$ to $66.04\%$ and linear-weighted Cohen's $\kappa$ from $0.518$ to $0.538$. Exact agreement measures matching labels; weighted $\kappa$ accounts for chance agreement and the order \textsc{Fail}/\textsc{Partial}/\textsc{Pass}. Main Table~\ref*{tab:judge_proxy_agreement} reports both metrics overall and by track.}

\paragraph{Scoring conventions.}
\auditfix{Strict Completion is the fraction of nonempty implementation criteria receiving a final \textsc{Pass}. Graded Completion uses the evaluator's recorded earned and maximum points, including half credit for \textsc{Partial}. Both are reported on a $0$--$100$ scale (Section~\ref*{sec:eval:scoring}). Task means use equal weights over the fixed model--task grid, with failed evaluations contributing zero; criterion-verdict shares pool individual judgments. Non-editing conversational and diagnostic auxiliary records are excluded from scoring denominators. Seven \tracktwo{} tasks have differing recorded criterion counts across model runs; the completion-gap analysis retains each run's recorded denominator (Section~\ref{sec:app:gap}).}

\paragraph{Evidence requirements by criterion type.}
Reachable source and DOM state can support static criteria. Visual claims require a successful render and screenshot evidence; interaction claims also require recorded user actions and the resulting state transition. Internal state or a handler in source code cannot substitute for an executed interaction. Failed builds or missing DOM nodes invalidate render-dependent claims, and missing required evidence cannot be replaced with a weaker proxy.

\paragraph{\auditfix{Execution and evidence retention.}}
\auditfix{The candidate runs in an isolated workspace. Fixed recovery heuristics address non-blocking environment issues before the \texttt{vite build} or static HTML pipeline, server startup, and headless Chrome rendering. A shared evaluator configuration and Playwright automation collect source diffs, DOM trees, body text, screenshots, and source-to-entry reachability logs; interaction checks also record action sequences and network telemetry. The evaluator applies the hidden rubric to this evidence with the baseline available for comparison. Available archived records link requests and rubrics to retained workspaces, build, server, and browser outcomes, evidence, and verdicts. Section~\ref{sec:app:failure_traj_cases} describes the retained evidence for the case studies.}

\subsection{Conditional Failure Attribution}
\label{sec:app:failure_attribution}

Attribution is triggered for low-scoring cases ($G_t<0.8$) when an independently rendered baseline passes the startability gate. A parallel solvability probe attempts a minimal patch in an isolated sandbox, stubbing inaccessible network endpoints while preserving core business logic. Unfulfilled criteria are attributed to model limitations, environmental blockage, mixed causes, or initial-artifact defects. These diagnoses leave raw scores and denominators unchanged. \auditfix{Under the attribution rule in Section~\ref*{sec:eval:scoring}, any supplementary adjusted score may grant credit only for established infrastructure failures.}


\FloatBarrier
\section{Task-Relevance Questionnaire}
\label{sec:app:questionnaire}

\auditfix{In this LLM-simulated pilot, Co-Design receives the highest mean ratings for perceived real-world fit and practical value, and ranks second for request frequency and wording naturalness (Table~\ref{tab:task_relevance_pilot}). DesignBench-edit leads on the latter two dimensions. Averaging the four dimensions equally, Co-Design scores highest ($4.31$), followed by DesignBench-edit ($3.93$), WebCompass-edit ($3.53$), and MiniAppBench ($3.33$). These model-generated ratings describe the sampled requests under the simulated occupational profiles.}

\begin{table}[!htbp]
\centering
\caption{\textbf{LLM-reported task-relevance ratings.} GPT-6~Astra generates ratings for $42$ fictional occupational profiles, each applied to the same $20$ tasks, with five tasks per benchmark. No human respondents participate. Scores range from $1$ to $5$. Fit = perceived real-world fit; Value = practical value; Freq. = simulated request frequency; Nat. = wording naturalness. Average is the unweighted mean of the four displayed dimension scores; bold marks column maxima.}
\label{tab:task_relevance_pilot}
\footnotesize
\setlength{\tabcolsep}{4pt}
\renewcommand{\arraystretch}{1.10}
\begin{tabular*}{\linewidth}{@{\extracolsep{\fill}}lrrrrr@{}}
\toprule
\cdhead{Benchmark} & \cdhead{Fit} & \cdhead{Value} & \cdhead{Freq.} & \cdhead{Nat.} & \cdhead{Average} \\
\midrule
WebCompass-edit & 4.00 & 4.06 & 2.77 & 3.30 & 3.53 \\
Co-Design & \cdbest{4.65} & \cdbest{4.78} & 3.36 & 4.46 & \cdbest{4.31} \\
DesignBench-edit & 4.60 & 2.80 & \cdbest{3.69} & \cdbest{4.64} & 3.93 \\
MiniAppBench & 4.18 & 3.63 & 1.58 & 3.94 & 3.33 \\
\bottomrule
\end{tabular*}
\end{table}

\paragraph{Simulated profiles and tasks.}
GPT-6~Astra generates all questionnaire responses under $42$ fictional profiles spanning seven occupational roles. Each profile is used to rate $20$ tasks, giving $840$ model-generated task-level responses. No human respondents participate in this pilot. We randomly sample five tasks from each of four benchmarks: WebCompass-edit, Co-Design, DesignBench-edit, and MiniAppBench. The Co-Design sample includes tasks from both the \trackone{} and \tracktwo{} tracks. All profiles are used to evaluate the same $20$ requests. Table~\ref{tab:questionnaire_roles} lists the seven occupational roles used in the simulation.

\begin{table}[!htbp]
\centering
\caption{\textbf{Occupational roles used in the LLM-simulated questionnaire.} The $42$ fictional profiles span seven occupational roles; all responses are model-generated.}
\label{tab:questionnaire_roles}
\small
\begin{tabular*}{\linewidth}{@{\extracolsep{\fill}}p{0.94\linewidth}@{}}
\toprule
\cdhead{Simulated occupational role} \\
\midrule
Frontend developer \\
Full-stack developer \\
Product manager \\
UI/UX designer \\
Software tester \\
Independent app builder \\
Business operations / internal-tool user \\
\bottomrule
\end{tabular*}
\end{table}

\cdneedspace{8\baselineskip}
\paragraph{Questionnaire dimensions.}
The questionnaire evaluates four dimensions on a five-point scale, with ratings generated from the perspective of each fictional profile:

\begin{enumerate}[leftmargin=*,itemsep=2pt,topsep=3pt]
\item \textbf{Fit, perceived real-world fit:} How closely the request fits real development use cases, from $1$ (not at all) to $5$ (very closely).
\item \textbf{Value, practical value:} The practical benefit of completing the request for users, from $1$ (almost none) to $5$ (very high).
\item \textbf{Freq., request frequency:} How often someone in the fictional profile's occupational role might encounter similar requests, from $1$ (never) to $5$ (very often).
\item \textbf{Nat., wording naturalness:} How similar the wording is to a request made during actual collaboration, from $1$ (not at all similar) to $5$ (very similar).
\end{enumerate}

\paragraph{Aggregation.}
For each dimension, we first average the available model-generated profile ratings within each task, then equally average the five task means for each benchmark. The Average column reports the unweighted mean of the four displayed dimension scores.

\paragraph{Scope and limitations.}
\auditfix{All $840$ task-level responses and ratings in Table~\ref{tab:task_relevance_pilot} are generated by GPT-6~Astra. The fictional profiles are role-conditioning contexts, rather than independent human respondents. Results depend on the model, prompts, and the $20$ sampled requests. They summarize simulated impressions and cannot establish actual human preferences, real-world request frequencies, or the representativeness of occupational groups.}

\FloatBarrier
\begin{table}[!tp]
\centering
\caption{\textbf{Failure types in $200$ randomly sampled incomplete runs.} A miss can carry several types, and a run counts toward a type if at least one of its misses does, so shares do not sum to $100\%$. The \trackone{} and \tracktwo{} columns give runs and shares of the $83$ and $117$ sampled runs; Pooled weights the two tracks by their $765$ and $1{,}566$ current incomplete runs and gives MOVER 95\% intervals. Rows under \emph{No functional error shown} count only misses that carry no agent-error type. Retained annotations predate refinements to the logic-error and build-break definitions. LLM coding agents, including GPT-6~Astra (itself an evaluated model), assigned the labels; no human verified them.}
\label{tab:failure_types}
\footnotesize
\setlength{\tabcolsep}{3pt}
\renewcommand{\arraystretch}{1.04}
\begin{tabular*}{\linewidth}{@{\extracolsep{\fill}}>{\raggedright\arraybackslash}p{0.19\linewidth}>{\raggedright\arraybackslash}p{0.415\linewidth}rrrrrl@{}}
\toprule
& & \multicolumn{2}{c}{\cdhead{\trackone{} ($n{=}83$)}} & \multicolumn{2}{c}{\cdhead{\tracktwo{} ($n{=}117$)}} & \multicolumn{2}{c}{\cdhead{Pooled}} \\
\cmidrule(lr){3-4}\cmidrule(lr){5-6}\cmidrule(l){7-8}
\cdhead{Type} & \cdhead{Definition} & \cdhead{Runs} & \cdhead{\%} & \cdhead{Runs} & \cdhead{\%} & \cdhead{\%} & \cdhead{95\% CI} \\
\midrule
\multicolumn{8}{@{}l}{\textit{Agent errors}} \\
Misread content or reference & Wrong audio or video recognition, text extraction, translation, fact, meaning, or target content; a style mismatch alone does not count. & 61 & 73.5 & 6 & 5.1 & 27.6 & [23.7, 32.2] \\
Edited the wrong target & The evidence shows that the wrong element, region, or object was edited and the target was not; missing one target alone does not count. & 8 & 9.6 & 3 & 2.6 & 4.9 & [3.0, 9.0] \\
Visual or layout mismatch & Color, size, layout, font, motion, or responsive rendering of the target does not match. & 15 & 18.1 & 35 & 29.9 & 26.0 & [20.5, 32.7] \\
Missing feature or interaction & A requested control, flow, event handler, entry point, or data wiring is missing or only a placeholder. & 6 & 7.2 & 51 & 43.6 & 31.7 & [25.7, 38.2] \\
Logic or data error & Wrong algorithm, calculation, validation, parsing, formatting, boundary handling, or business rule. & 4 & 4.8 & 27 & 23.1 & 17.1 & [12.5, 23.2] \\
Build or runtime break & The output does not build or run (syntax, types, dependency wiring, exceptions); if the edit is not shown to cause it, the miss is also labeled unresolved. & 0 & 0.0 & 1 & 0.9 & 0.6 & [0.1, 3.5] \\
Regression or side effect & Before--after evidence shows that non-target content, appearance, or behavior was broken or changed against a preservation requirement. & 12 & 14.5 & 6 & 5.1 & 8.2 & [5.5, 13.0] \\
Missed dependent update & A change to configuration, an interface, code, copy, or a data representation misses known dependent places. & 1 & 1.2 & 8 & 6.8 & 5.0 & [2.7, 9.4] \\
Runtime state not synced & State is not updated, or is wrongly propagated, between inputs, selections, asynchronous results, caches, storage, URLs, or linked views. & 0 & 0.0 & 32 & 27.4 & 18.4 & [13.5, 24.4] \\
\midrule
\multicolumn{8}{@{}l}{\textit{No functional error shown}} \\
Implementation form only & A criterion requires a specific class name, DOM structure, implementation method, file location, or no unrelated additions, and no functional failure is shown. & 1 & 1.2 & 50 & 42.7 & 29.1 & [23.3, 35.4] \\
Environment or evaluation issue & Evidence of an external environment block, evaluator rewriting, unusable test input, or conflicting requirements; not a confirmed agent error. & 2 & 2.4 & 29 & 24.8 & 17.4 & [12.7, 23.5] \\
Unresolved evidence & The report is missing or blank or does not describe the key failure verifiably; a review status. & 0 & 0.0 & 10 & 8.5 & 5.7 & [3.2, 10.3] \\
\midrule
\multicolumn{8}{@{}l}{\textit{Combinations}} \\
Regression or missed update & At least one regression or side effect, or missed dependent update. & 13 & 15.7 & 14 & 12.0 & 13.2 & [9.4, 18.9] \\
\ldots{} or runtime state & As above, or at least one runtime state not synced. & 13 & 15.7 & 43 & 36.8 & 29.8 & [24.0, 36.6] \\
No agent-error type & No miss carries an agent-error type. & 1 & 1.2 & 20 & 17.1 & 11.9 & [8.0, 17.4] \\
\bottomrule
\end{tabular*}
\end{table}

\makeatletter
\renewcommand\paragraph{\@startsection{paragraph}{4}{\z@}{6pt plus 1pt minus 1pt}{-1em}{\normalfont\normalsize\bfseries}}
\makeatother
\section{Failure Types and Trajectories}
\label{sec:app:failure_traj}

\subsection{Failure-Type Sample}
\label{sec:app:failure_types}

\auditfix{In the $200$-run sample, misreading content or references is the most common agent-error label on \trackone{}; on \tracktwo{}, missing features or interactions, visual mismatches, and unsynchronized runtime state are the leading agent-error labels. Table~\ref{tab:failure_types} gives the definitions, counts, and intervals underlying Figure~\ref*{fig:failure_types}a and Section~\ref*{sec:exp:failures} of the main paper.}

\paragraph{Findings.}
Every miss carries an agent-error type in $80$ of the $83$ \trackone{} runs but only $48$ of the $117$ \tracktwo{} runs. In $20$ \tracktwo{} runs and one \trackone{} run, no miss does: eight of these \tracktwo{} runs fail only on implementation form, nine only on environment or evaluation issues or unresolved misses, and three on a mix of both. These runs remain failures in every reported score; no run was re-scored for attribution. We checked the implementation-form criterion texts against the full requests by string matching: $38$ of the $136$ \tracktwo{} misses labeled implementation form quote identifiers (test identifiers, class names, or variable names), and $38$ of the $40$ identifier mentions (counted per run) do not occur in the request; none of the $88$ distinct \tracktwo{} requests in the sample mentions a test identifier.

\paragraph{\auditfix{Sample composition.}}
The population is the $2{,}331$ current incomplete runs of the ten models in Table~\ref*{tab:leaderboards}: $765$ on \trackone{} and $1{,}566$ on \tracktwo{}. Four zero-score runs lack criterion records and remain in the population; none was drawn. The sample has $83$ \trackone{} and $117$ \tracktwo{} runs on $144$ tasks, covers all ten models, and contains $713$ misses ($225$ and $488$); \trackone{} is over-represented by chance ($66$ runs expected), which the pooled weights correct.

\paragraph{\auditfix{Counting conventions.}}
Every miss receives one or more types; a run carries the union of its misses' types, and $104$ of the $200$ runs carry more than one label. A type therefore marks evidence that a run shows, not the single cause of its failure. The three grey types of Figure~\ref*{fig:failure_types} count only misses that carry no agent-error type; counting every miss with a grey label instead, including the $24$ \tracktwo{} misses ($28$ in both tracks) that also carry an agent-error type, gives $47.9\%$, $25.6\%$, and $10.3\%$ on \tracktwo{}. One retained run has a build break, all eight misses also unresolved; it stays among runs with an agent-error type because its output does not run. Applying the build-break definition's responsibility caveat instead would give $21$ \tracktwo{} runs ($18\%$) without an agent-error type in this analysis.

\paragraph{Estimation.}
Track shares are sample proportions with Wilson 95\% intervals. Pooled shares weight the tracks by their current incomplete runs ($765/2{,}331$ and $1{,}566/2{,}331$) and combine the per-track Wilson limits with the MOVER method; the intervals omit the finite-population correction and do not reflect labeling error.

\paragraph{\auditfix{Sampling procedure.}}
We maintain a simple random sample of $200$ runs using a label-independent refresh (Python \texttt{random.Random}, initial seed $20261107$, refresh seed $20261111$): the number of common and new-only identities follows its exact hypergeometric law, the common subset is uniformly subsampled or extended from the initial uniform draw, and new-only identities are drawn uniformly. This preserves uniform sampling without replacement from the current population. The refresh retains the same $200$ model--task identities; every sampled Grok trajectory is replaced by its current scored run.

\paragraph{\auditfix{Annotation procedure.}}
GPT-6~Astra at maximum reasoning effort in Codex assigned the retained annotations on $172$ unchanged scored runs. A Codex LLM coding agent read the full requests, criterion texts, and current final judge evidence and assigned all $84$ miss labels on the $28$ current Grok trajectories \auditfix{using the same twelve failure-type categories.} No human reviewed the codebook or labels, and neither labeling pass was blinded. For retained annotations, the labeller typically saw only the first $200$ characters of the request and the first $60$ of each criterion, along with the judge's evidence; it could also see prior targeted-screen labels. The logic-error and build-break definitions were refined after the initial pass; retained labels were not revisited, while the new annotations use the refined definitions. Because the retained annotator is also evaluated, we note that $5$ of its $10$ sampled \tracktwo{} runs carry no agent-error type, against $15$ of the other models' $107$ (two-sided Fisher exact $p=0.01$). Both self-preference and stronger models' remaining failures being more often form-only or evaluation-side would produce this pattern; the sample cannot separate them.

\subsection{Failure Trajectories}
\label{sec:app:failure_traj_cases}

\paragraph{Common pattern.}
In these trajectories, the agents' checks test the interpretation they have already chosen, through their own crops, renders, or tests, rather than the alternative that would expose the error. Their final reports state the result without qualification. \auditfix{Figure~\ref*{fig:failure_modes}b--e presents four final-output examples; Figure~\ref{fig:failure_traj} traces the operations behind three of them.}

\auditfix{The distinction is between confirming an observed result and checking it against the request. A crop can confirm a visual change while dropping the identifier that selects the target; a successful render can confirm a replacement without establishing which assets required it. Likewise, testing two actions separately does not establish their behavior when combined. These cases trace the gap between the evidence gathered and the requirement being tested.}


\paragraph{Case selection.}
We list criteria on both tracks that at least six of the original ten endpoints fail and inspect each candidate's request, attachments, final workspaces, evaluator evidence, and solver sessions. We drop candidates whose request admits another reasonable reading, whose criterion asks for more than the request states, or whose failure cannot be confirmed from the archived workspace. Figure~\ref*{fig:failure_modes}b--e in the main paper shows four cases that remain, three from \trackone{} and one from \tracktwo{}; Figure~\ref{fig:failure_traj} links three of them to the operations that produce them.

\paragraph{Misread content or reference.}
No frame shows the bus number in full: the bus's mirror covers the leading 1 until about $57$\,s and the 3 from about $59$\,s. \texttt{Qwen3.8-27B} extracts frames from $44$\,s on and opens two in which the 1 is covered. It notes that ``the `1' is partially cut off on the left'' and concludes ``This confirms the body number is `13'\,''. It had planned to open a frame at $59$\,s, where the leading 1 is visible, but never does. \texttt{kimi-k3} settles on the ``13'' it sees at $42$\,s. The successful \texttt{opus-4-8} writes that the number ``may have a leading digit'', widens its crop, ranks frames from $39$ to $47$\,s by sharpness, and reads 113. Two of the nine failures have other causes: \texttt{grok-4.6} answers from frames taken before the bus appears, and \texttt{GLM-5.2} reports OCR noise.

\begin{minipage}{\columnwidth}
\centering
\captionsetup{skip=4pt,hypcap=false}
\includegraphics[trim=0bp 0bp 480bp 0bp,clip,width=0.55\linewidth]{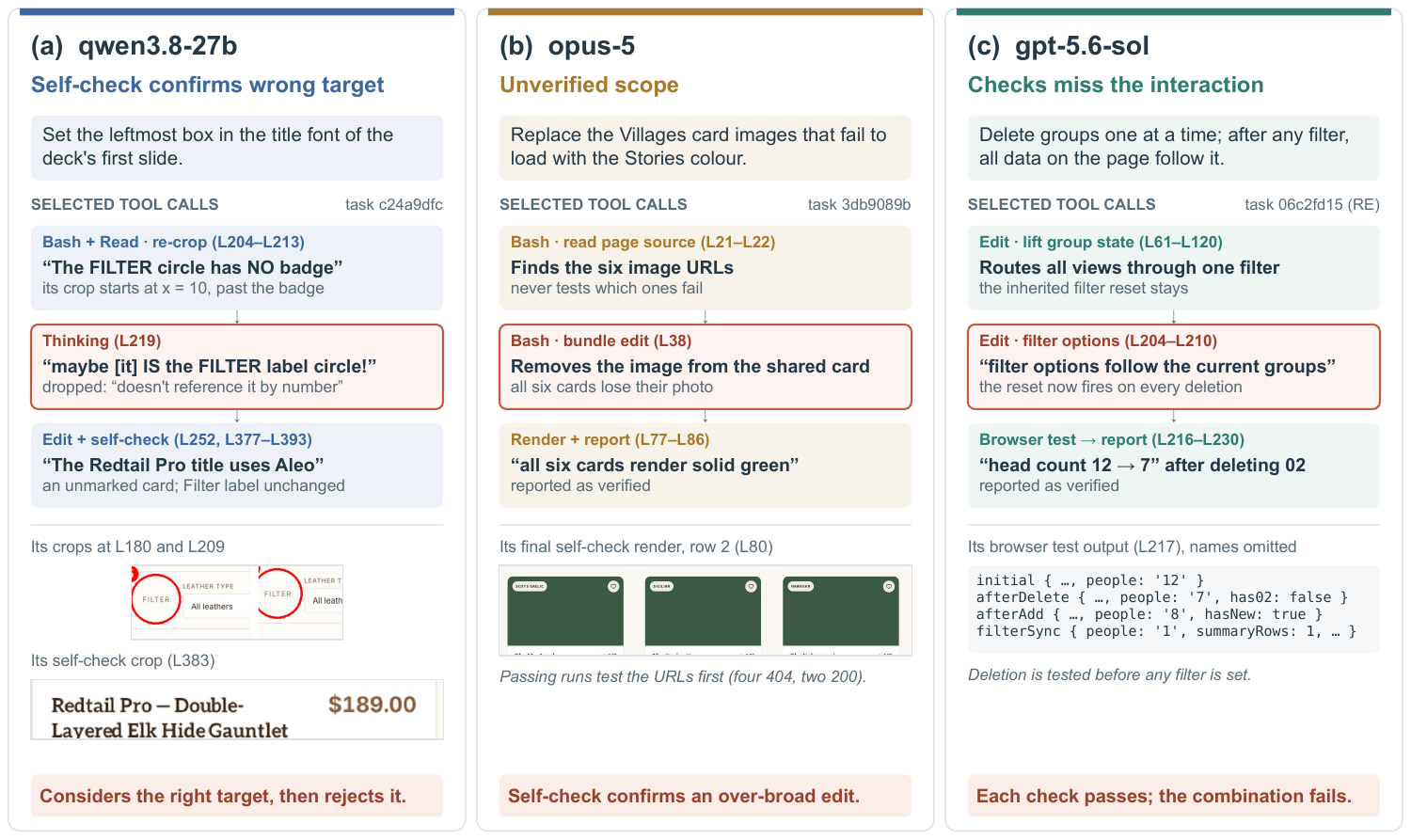}
\captionof{figure}{\textbf{Selected operations in three failure trajectories.} Line numbers refer to the archived solver sessions, and the outlined step is decisive. (a)~\texttt{Qwen3.8-27B} crops away the badge of mark~5, considers the Filter label as the target, and rejects it.}
\label{fig:failure_traj}
\end{minipage}
\par\vspace{6pt}

\paragraph{Edited the wrong target.}
\texttt{Qwen3.8-27B} reads the clipped badge of mark~5 as ``2'', re-crops the mark from $x{=}10$, which cuts the badge away, and concludes that ``the FILTER circle has NO badge''. It then considers the correct target, ``maybe [the leftmost box] IS the `FILTER' label circle!'', but sets it aside because ``the task doesn't reference it by number''. It restyles the unmarked Redtail Pro card and confirms the change on a crop of that card's title (Figure~\ref{fig:failure_traj}a). The five other failing endpoints restyle the card under mark~3, which the time-swap instruction already uses. \auditfix{Once the badge is cropped away, that crop no longer preserves the numbered marker used to select the target. Checking the title of the chosen card confirms its appearance after the edit, but does not check whether the request selected that card. The target-selection error thus survives the visual self-check.}

\cdwidepage{\centering
\includegraphics[trim=240bp 0bp 240bp 0bp,clip,width=0.44\linewidth]{figures/failure_trajectories.pdf}\hspace{0.04\linewidth}%
\includegraphics[trim=480bp 0bp 0bp 0bp,clip,width=0.44\linewidth]{figures/failure_trajectories.pdf}
\captionsetup{type=figure}\ContinuedFloat
\caption{\textbf{Selected operations in three failure trajectories (continued).} (b)~\texttt{opus-5} replaces every card image without testing which ones fail and reports the result as verified. (c)~\texttt{gpt-5.6-sol} makes the inherited filter reset fire on every deletion and tests deletion and filtering only one after the other. Images and the test log are those the agents viewed; quotes in (c) are translated from Chinese, and names are omitted from its log.}
}

\paragraph{Regression or side effect.}
\texttt{opus-5} reads the six image URLs from the source map but never requests them. It removes the image element from the shared card component, renders the page only after this edit, and reports that ``all six cards render solid green media panels'' (Figure~\ref{fig:failure_traj}b). All four passing endpoints request the URLs before editing, finding four that return 404 and two that return 200, and limit the change to the failing images; three of them use an error handler on the image.

\paragraph{Runtime state not synced.}
The starting app of this \tracktwo{} task resets every filter to all values whenever the statistics change (\texttt{StatsTable.tsx}, lines 68--74). \texttt{gpt-5.6-sol} lifts the group state so that one filtered data chain drives all views.
\par\noindent In a browser it then checks that deleting group~02 lowers the head count from 12 to 7 and, afterwards, that a name filter leaves one person. When this check shows that the filter still lists deleted members, it makes the filter options, and with them the inherited reset, depend on the groups. It never deletes a group while a filter is set, and it reports both checks as passing (Figure~\ref{fig:failure_traj}c). All six failing endpoints keep the reset; \texttt{opus-4-8} and \texttt{Qwen3.8-27B} add the groups to its dependencies. 
Among the passing endpoints, \texttt{GLM-5.2} keeps the previous selection when the values change, and \texttt{opus-5} and \texttt{glm-5.3-flash} apply the group scope and the filter in one pipeline; the fourth, \texttt{gpt-6-astra}, has no archived workspace.

\FloatBarrier
\cdneedspace{18\baselineskip}
\section*{LLM Usage Statement}
\label{sec:llm_usage}

LLMs were used as the evaluated coding agents and as components of the evaluation protocol. Claude Opus~4.8 served as the primary judge and refuter; GPT-6~Astra and Gemini~3.1~Pro were alternative judges. The evaluated solver endpoints and agent harnesses are identified in Section~\ref{sec:exp:setup} and Appendix~\ref*{sec:app:cross_harness}. GPT-6~Astra generated all $840$ task-level questionnaire responses under $42$ fictional occupational profiles, with no human respondents, and performed failure-type coding in Codex; a subsequent Codex coding agent updated the Grok labels. The procedures and limitations are described in Appendix~\ref*{sec:app:questionnaire} and~\ref*{sec:app:failure_types}.

OpenAI Codex assisted with manuscript drafting, revision and language editing; analysis and figure-generation code; case-study figures and captions; and LaTeX formatting. It also assisted with statistical analyses, archived-run audits, and failure-case selection and interpretation. LLM coding agents using the tools named in this statement also assisted with starting-application repair during benchmark construction (Appendix~\ref*{sec:app:curation}). The authors take responsibility for the submitted content, including scientific claims, results, citations, code, and figures.

\end{document}